\documentclass{article}

\usepackage[utf8]{inputenc} 
\usepackage[T1]{fontenc}    
\usepackage{hyperref}       
\usepackage{url}            
\usepackage{booktabs}       
\usepackage{amsfonts}       
\usepackage{nicefrac}       
\usepackage{microtype}      

\usepackage{iclr2019_conference,times}

\usepackage[bottom]{footmisc}
\usepackage{amsmath}
\usepackage{todonotes}
\usepackage{subcaption}
\usepackage{wrapfig}
\usepackage{bm}
\usepackage{url}
\usepackage{booktabs}       
\usepackage{amsfonts}       

\usepackage{varioref}
\definecolor{mydarkblue}{rgb}{0,0.08,0.45}
\usepackage{hyperref}
\hypersetup{ %
   breaklinks=true,
   unicode=true,
   pdftitle={},
   pdfauthor={},
   pdfsubject={},
   pdfkeywords={},
   pdfborder=0 0 0,
   pdfpagemode=UseNone,
   colorlinks=true,
   linkcolor=mydarkblue,
   citecolor=mydarkblue,
   filecolor=mydarkblue,
   urlcolor=mydarkblue,
   pdfview=FitH}
\usepackage[all]{hypcap}
\usepackage{cleveref}

\iclrfinalcopy

\title{Investigating Object Compositionality in \\ Generative Adversarial Networks\thanks{A preliminary version~\citep{steenkiste2018case} of this work appeared under a different title as a workshop paper at NeurIPS 2018 and as a technical report on arXiv (v1).}}

\author{
  Sjoerd van Steenkiste\thanks{The majority of this work was done while at Google Brain. Correspondence to \texttt{sjoerd@idsia.ch}.} $^{,1}$, 
  Karol Kurach$^{2}$,
  J\"urgen Schmidhuber$^{1,3}$,
  Sylvain Gelly$^{2}$ \\
  $^1$ IDSIA, USI, SUPSI, $^2$ Google Brain, $^3$ NNAISENSE
  \\\And 
}

\begin{document}

\maketitle

\begin{abstract}
Deep generative models seek to recover the process with which the observed data was generated.
They may be used to synthesize new samples or to subsequently extract representations.
Successful approaches in the domain of images are driven by several core inductive biases.
However, a bias to account for the compositional way in which humans structure a visual scene in terms of objects has frequently been overlooked. 
In this work, we investigate object compositionality as an inductive bias for Generative Adversarial Networks (GANs).
We present a minimal modification of a standard generator to incorporate this inductive bias and find that it reliably learns to generate images as compositions of objects. 
Using this general design as a backbone, we then propose two useful extensions to incorporate dependencies among objects and background.
We extensively evaluate our approach on several multi-object image datasets and highlight the merits of incorporating structure for representation learning purposes.
In particular, we find that our structured GANs are better at generating multi-object images that are more faithful to the reference distribution.
More so, we demonstrate how, by leveraging the structure of the learned generative process, one can `invert' the learned generative model to perform unsupervised instance segmentation.
On the challenging CLEVR dataset, it is shown how our approach is able to improve over other recent purely unsupervised object-centric approaches to image generation.
\end{abstract}

\section{Introduction}
\begin{figure}[b]
\centering
\includegraphics[width=\linewidth]{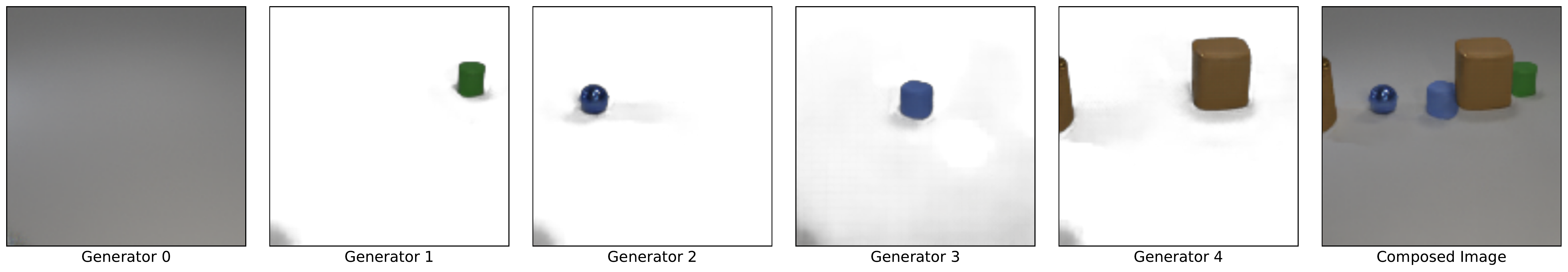}
\caption{In this paper, we investigate three simple modifications to a neural network generator that allows it to generate images as a composition of a generated background (left) and generated individual objects (middle). The resulting generated image is shown on the right.}
\label{fig:clevr_split}
\end{figure}

Probabilistic generative models aim to recover the process with which the observed data was generated.
It is postulated that knowledge about the generative process exposes important factors of variation in the environment (modeled by latent variables) that may subsequently be obtained using an appropriate posterior inference procedure.
Hence, in order to use generative models for representation learning, it is important that the \emph{structure} of the generative model closely resembles the underlying generative process. 

Deep generative models of images rely on the expressiveness of neural networks to learn the generative process directly from data~\citep{goodfellow2014generative,kingma2014stochastic,oord2016pixel}.
Their structure is determined by the \emph{inductive bias} of the neural network, which steers it to organize its computation in a way that allows salient features to be recovered and ultimately captured in a representation~\citep{kingma2014stochastic,dinh2017density,donahue2017adversarial,dumoulin2017adversarially}.
Recently, it was shown that independent factors of variation, such as pose and lighting of human faces may be recovered in this way~\citep{chen2016infogan,higgins2017beta}.

A promising, but under-explored, inductive bias in deep generative models of images is \emph{compositionality at the representational level of objects}, which accounts for the compositional nature of the visual world and our perception thereof~\citep{spelke2007core,battaglia2013simulation}.
It allows a generative model to describe a scene as a composition of objects, thereby `disentangling' visual information that can be processed largely independently of one another.
By explicitly accounting for object compositionality one imposes an \emph{invariance} that simplifies the generative process, and potentially allows one to (more efficiently) learn a more accurate generative model of images.
Moreover, by explicitly considering individual objects at a representational level, (posterior) inference in the learned generative model allows one to recover corresponding object representations that are desirable for many `down-stream' learning tasks~\citep{santoro2017simple,janner2018reasoning}.\looseness=-1

In this work, we investigate object compositionality for Generative Adversarial Networks (GANs)~\citep{goodfellow2014generative}\footnote{There exist interesting parallels between GANs and earlier approaches~\citep{schmidhuber1990making,schmidhuber1992learning}, and we refer the reader to \cite{schmidhuber2020generative} for a comparison.}.
We investigate a minimal modification to a standard neural network generator that incorporates the right inductive bias and find that it reliably learns to generate images as compositions of objects. 
Using this general design as a backbone, we then propose two useful extensions that provide a means to incorporate dependencies among objects and background in the generative process.
Finally, we demonstrate how one can leverage the learned structured generative process, which is now interpretable and semantically understood, to perform inference and recover individual objects from multi-object images without additional supervision.

The goal of this work is not to improve upon state-of-the-art approaches to multi-object image generation that rely on prior knowledge by \emph{conditioning} on explicit scene graphs~\citep{johnson2018image,xu2018deep}, images of individual objects~\citep{lin2018st,azadi2019compositional}, or their properties~\citep{hinz2018generating}.
Rather, we suggest an alternative approach to learn deep generative models for complex visual scenes based on GANs, that does not rely on prior knowledge in the form of conditioning.
Instead, our approach \emph{learns} to separate information about objects at a representation level, so that it can be `inverted' to extract this knowledge.
As a consequence, without relying on prior knowledge in the form of auxiliary inputs that are generally not available, the approach presented here contributes an important step in developing purely unsupervised generative models for representation learning purposes that are expected to be more applicable in the context of AI~\citep{bengio2013representation}.

We extensively evaluate our structured generator\footnote{Code is available online at \url{https://git.io/JePuK}.} on several multi-object image datasets.
When using our inductive bias, we find that GANs are able to learn about the individual objects and the background of a scene, without prior access to this information (Figure~\ref{fig:clevr_split}).
The results of our quantitative experiments highlight the merits of incorporating these inductive biases for representation learning purposes.
In particular, compared to a strong baseline of popular GANs including recent state-of-the-art techniques, our approach consistently outperforms in generating better images that are more faithful to the reference distribution.
More so, we demonstrate how, by leveraging the structure of the learned generative process, one can perform inference in these models and perform unsupervised instance segmentation.
On the challenging CLEVR dataset~\citep{johnson2017clevr} it is shown how our approach is able to improve over other recent purely unsupervised object-centric approaches to image generation.

\section{Generative Adversarial Networks}
Generative Adversarial Networks (GANs) are powerful generative models that learn a stochastic procedure to generate samples from a distribution $P(X)$.
Traditionally GANs consist of two deterministic functions: a generator $G(\bm{z})$ and a discriminator (or critic) $D(\bm{x})$.
The goal is to find a generator that accurately transforms samples from a prior distribution $\bm{z}\sim P(Z)$ to match samples from the target distribution $\bm{x}\sim P(X)$.
This can be done by using the discriminator to implement a suitable objective for the generator, in which it should behave \emph{adversarially} with respect to the goal of the discriminator in determining whether samples $\bm{x}$ were sampled from $P(X)$ or $G(P(Z))$ respectively.
These objectives can be summarized as a \emph{minimax} game with the following value function:

\begin{align}
\label{eq:gan}
\min_G \max_D V(D, G) = \mathbb{E}_{\bm{x} \sim P(X)} \left[\log D(\bm{x})\right] + \mathbb{E}_{\bm{z}\sim P(Z)} \left[\log(1 - D(G(\bm{z}))) \right].
\end{align}

When the generator and the discriminator are implemented with neural networks, optimization may proceed through alternating (stochastic) gradient descent updates of their parameters with respect to \eqref{eq:gan}. However, in practice, this procedure might be unstable and the minimax formulation is known to be hard to optimize.
Many alternative formulations have been proposed and we refer the reader to \cite{lucic2018gans} and \cite{kurach2019large} for a comparison.

Based on the findings of \cite{kurach2019large} we consider two practical reformulations of \eqref{eq:gan} in this paper: Non-Saturating GAN (NS-GAN)~\citep{goodfellow2014generative}, in which the generator maximizes the probability of generated samples being real, and Wassertein GAN (WGAN)~\citep{arjovsky2017wasserstein} in which the discriminator minimizes the Wasserstein distance between $G(P(Z))$ and $P(X)$.
In both cases, we explore two additional techniques that have proven to work best on a variety of datasets and architectures: the gradient penalty from \cite{gulrajani2017improved} to regularize the discriminator, and spectral normalization~\citep{miyato2018spectral} to normalize its gradients.

\section{Incorporating Structure}
In order to formulate the \emph{structure} (inductive bias) required to achieve object compositionality in neural network generators, we will focus on the corresponding type of invariance that we are interested in.
It is concerned with independently varying the different objects that an image is composed of, which requires these to be identified at a representational level and described in a common format.

In the following subsections, we first present a simple modification to a neural network generator to obtain object compositionality.
Afterward, we present two useful extensions that allow one to additionally reason about relations between objects, and to model background and occlusions.

\begin{figure*}[t]
    \centering
    \includegraphics[width=\linewidth]{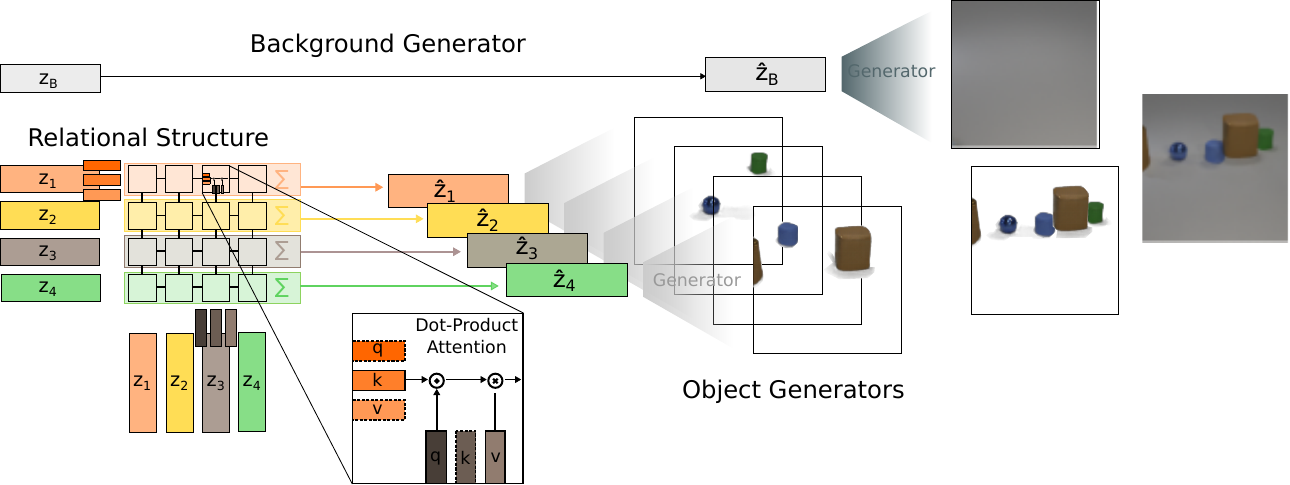}
    \caption{We propose three modifications to a standard neural network generator to generate images as a composition of individual objects and background. In this case, it consists of $K=4$ \emph{object generators} (shared weights) that each generate an image from separate latent vector $\bm{z}_i \sim P(Z)$, which serve as the backbone for object compositionality. On the left side, a \emph{relational stage} is shown as one possible extension to model relations between objects by first computing $\bm{\hat{z}}_i$ from $\bm{z}_i$ that are then fed to the object generators. In the back, a second extension is shown that incorporates a \emph{background generator} (unique weights) to generate a background image from a separate latent vector $\bm{z}_b \sim P(Z_b)$. The whole system is trained end-to-end as in the standard GAN framework, and the final image is obtained by composing (in this case using alpha compositing) the outputs of all generators.}
    \label{fig:drawing}
\end{figure*}

\subsection{Object Compositionality}
\label{section:strict_independence}

A minimal modification of a neural network generator assumes that images $\bm{x}\sim P(X)$ are composed of objects that are independent of one another. 
For images having $K$ objects, we consider $K$ i.i.d. vector-valued random variables $Z_{i}$ that each describe an object at a representational level.
$K$ copies of a deterministic generator $G(\bm{z})$ transform samples from each $Z_{i}$ into images, such that their superposition results in the corresponding output image:

\begin{equation}
\bm{x} = \sum_{i=1}^K G(\bm{z}_i), \ \text{where} \ \bm{z}_i \sim P(Z).
\label{eq:generator}
\end{equation}

When each copy of $G$ generates an image of a single object, the resulting generative model efficiently generates images in a compositional manner (Figure~\ref{fig:mm-split}).
Each object in \eqref{eq:generator} is described in terms of the same features (i.e. the $Z_i$'s are i.i.d) and the weights among the generators are shared, such that any acquired knowledge in generating a specific object is transferred across all others.
Hence, rather than having to learn about all combinations of objects (including their own variations) that may appear in an image, it suffices to learn about the different variations of each individual object.
This greatly simplifies the generative process without losing generality.

The superposition of generators in \eqref{eq:generator} can be viewed as a single generator of the form $G_{multi}(\bm{z} = [\bm{z}_1, \cdots, \bm{z}_K])$ and trained as before using the objective in \eqref{eq:gan}.
The generators in \eqref{eq:generator} do not interact, which prevents degenerate solutions and encourages $G$ to learn about objects.
However, it also implies that relations between objects \emph{within} a scene cannot be modelled in this way.
Finally, we note that the sum in \eqref{eq:generator} assumes that images only consist of objects, and that their values can be summed in pixel-space.
We will now present two extensions that focus on each of these aspects and incorporate additional structure besides the superposition of generators that serves as the basis for object compositionality.

\subsection{Relations Between Objects}
\label{section:relational_structure}

In the real world, objects are not strictly independent of one another.
Certain objects may only occur in the presence of others, or affect their visual appearance in subtle ways (eg. shadows).
In order to facilitate relationships of this kind our first extension consists of a \emph{relational stage}, in which the representation of an object is updated as a function of all others, before each generator proceeds to generate its image.

The relational stage is a \emph{graph network} which, in our case, consists of one or more self-attention blocks that compute these updates.
At the core of each attention block is Multi-Head Dot-Product Attention (MHDPA)~\citep{vaswani2017attention} that performs message-passing when one associates each object representation with a node in a graph~\citep{battaglia2018relational}.
This is a natural choice since it was shown that relations between objects and corresponding updates to their representations can be learned efficiently in this way \citep{zambaldi2018deep}.\looseness=-1

Similar to \cite{zambaldi2018deep}, a single head of an attention block first projects each latent vector $\bm{z}_i$ (associated with an object) into so-called query, key, and value vectors:\footnote{The role of the key and value vectors are analogous to those in a key-value database. However, in this case access takes place on the basis of similarity between key and query vectors (i.e. the attention weights), which allows this process to be differentiated.}

\begin{equation}
    \label{eq:attention_block-1}
    \bm{q}_i = \text{MLP}^{(\textit{q})}(\bm{z}_i), \ \ \bm{k}_i = \text{MLP}^{(\textit{k})}(\bm{z}_i), \ \ \bm{v}_i = \text{MLP}^{(\textit{v})}(\bm{z}_i).
\end{equation}

Next, the interaction of an object $i$ with all other objects (including itself) is computed as a weighted sum of their value vectors.
Weights are determined by computing dot-products between its query vector and all key vectors, followed by softmax normalization:

\begin{equation}
    \label{eq:attention_block-2}
    \bm{A} = \underbrace{softmax\big(\bm{Q}\bm{K}^{T} / \sqrt{d}\big)}_{\text{attention weights}}\bm{V},
\end{equation}

\noindent where $d=dim(\bm{v}_i)$ and $\bm{K},\bm{Q},\bm{V} \in \mathbb{R}^{K \times d}$ are matrices that contain the key, query, and value vectors for each object.
$\bm{A} \in \mathbb{R}^{K \times d}$ contains the result of each object attending to all other representations.  
Finally, we proceed by projecting each update vector $\bm{a}_i$ back to the original size of $\bm{z}_i$ before being added:

\begin{equation}
    \label{eq:attention_block-3}
     \bm{\hat{z}}_i = \text{MLP}^{\textit{(u)}}(\bm{a}_i) + \bm{z}_i.
\end{equation}

Additional heads (modeling different interactions) use different parameters for each MLP, and their outputs are combined with another MLP to arrive at a final $\bm{\hat{z}}_i$.
Complex relationships among objects can be modeled by using multiple attention blocks to \emph{iteratively} update $\bm{z}_i$.
A detailed overview of these computations can be found in \ref{app:experiment_details} and a schematic in Figure~\ref{fig:drawing}.

\subsection{Background and Occlusion}
\label{section:incorporating_background}

The second extension that we propose explicitly distinguishes the background from objects. 
Complex visual scenes often contain visual information that only appears in the background and does not occur frequently enough (or lacks a regular visual appearance) to be modeled as objects.
Indeed, one assumption that we have made is that objects can be compactly encoded in a representation $\bm{z}_i$ that is described in a \emph{common format}. 
Treating background as an ``extra'' object violates this assumption as the latent representations $\bm{z}_i$ (and corresponding generator) now need to describe objects that assume a regular visual appearance, as well as the background that is not regular in its visual appearance at all.
Therefore, we consider an additional \emph{background} generator (see Figure~\ref{fig:drawing}) having its own set of weights to generate the background from a separate vector of latent variables $\bm{z}_b \sim P(Z_b)$.
We will explore two different variations of this addition, one in which $z_b$ participates in the relational stage, and one in which it does not.

A remaining challenge is then in \emph{combining} objects with background and in modeling occlusion.
A straightforward adaptation of the sum in \eqref{eq:generator} to incorporate pixel-level weights would require the background generator to assign a weight of zero to all pixel locations where objects appear, thereby increasing the complexity of generating the background exponentially.
Instead, we require the object generators to generate an additional alpha channel for each pixel (while treating the output of the background generator as opaque), and use alpha compositing to combine the $K$ different outputs of the object generators $(\bm{x}_i, \bm{\alpha}_i)$ and background generator $(\bm{x_b}, 1)$ as follows:

\begin{equation}
    \bm{x} = \sum_{i=1}^{K} \left[ \bm{x}_i\bm{\alpha}_i \prod_{j=1}^{i-1} (1 - \bm{\alpha}_j) \right] + \bm{x}_b\prod_{i=1}^{K} (1 - \bm{\alpha}_i).
    \label{eq:alpha_composit}
\end{equation}

Using alpha compositing as in \eqref{eq:alpha_composit} is a standard technique in computer graphics, and therefore a natural choice in generating images.
On the other hand, alpha compositing uses a fixed ordering that assumes that the generated images (when overlapping) are provided in the correct order.
In principle, the relational stage can ensure that this is the case, although it may be difficult to learn in an adversarial setting.
An alternative choice would be to \emph{learn} to composit, for example by using a conditional GAN that starts from images of individual objects~\citep{lin2018st,azadi2019compositional}.
Here, we opt for the simpler choice of using \eqref{eq:alpha_composit} as we are primarily concerned with investigating the possibility of learning to generate images in this way.

\section{Related Work}
Prior works on incorporating inductive biases aimed at object compositionality typically treat an image as a \emph{spatial mixture} of image patches and utilizes multiple copies of the same function to arrive at a compositional solution.   
Different implementations use RBMs~\citep{le2011learning}, VAEs~\citep{nash17multi}, or (recurrent) auto-encoders inspired by EM-like inference procedures~\citep{greff2016tagger,greff2017neural} to generate these patches.
It was also shown that interactions between objects can be modeled efficiently in this way~\citep{steenkiste2018relational}.
Neither of these works has shown to be capable of generating more complex visual scenes that incorporate unstructured background as well as relations between objects.
By observing that GANs are often superior in terms of image generation capabilities, and by adding structure, we are able to improve upon these works in this regard.
Indeed, compared to the recent IODINE \citep{greff2019multi} that also utilizes a spatial mixture formulation, we will demonstrate how our approach better succeeds at modeling complex image distributions.

A conceptually similar line of related work uses variational inference to learn recurrent neural networks to \emph{iteratively} generate an image, one patch at a time~\citep{gregor2015draw,eslami2016attend,kosiorek2018sequential}.
The work by \cite{eslami2016attend} incorporates a strong inductive bias that associates each image patch with a single object.
However, also their approach is limited to generating (sequences) of binary images without background~\citep{eslami2016attend,kosiorek2018sequential}.
Related settings have also been explored with GANs using a recurrent generator~\citep{im2016generating,kwak2016generating}, while other work~\citep{yang2017lr} additionally considers a separate generator for the background that uses spatial transformations to integrate a foreground image.
From these, only the work of \cite{yang2017lr} briefly explores the problem of multi-object image generation on a dataset consisting of two non-overlapping MNIST digits.
Their approach is only moderately successful while making extra assumptions about the size of the digits.
Importantly, their approach does not provide a means to `invert' the learned model and perform inference. 

Other recent work in GANs has focused on \emph{conditional} image generation to simplify the task of multi-object image generation.
\cite{johnson2018image} generate multi-object images from explicit scene graphs, while \cite{xu2018deep} condition on a stochastic and-or graph instead.
\cite{azadi2019compositional} propose a framework to generate images composed of two objects by combining the images of each individual object.
Similarly, \cite{lin2018st} present an iterative scheme to remove or add objects to a scene based on prior knowledge about individual objects.
\cite{hinz2018generating} require object labels and bounding boxes to generate complex visual scenes consisting of multiple objects and background.
While these works generate realistic multi-object scenes, they require prior information (scene graphs, segmentations, etc.) about individual objects or scenes that is typically not available in many real-world settings.
Our work serves a complementary purpose in terms of image generation in that regard: while we consider visually simpler scenes, we do not require any extra information about scenes or objects.
In the context of representation learning, we are also able to `invert' the learned generative model by leveraging its structure to learn about objects.
This sets us firmly apart from these methods and also from standard unstructured GANs that do not rely on conditioning.
Indeed, without explicitly considering objects at a representational level (requiring architectural structure), one is unable to recover individual objects in this way.

In the context of unsupervised instance segmentation, three very recent works propose to structure the generator of a GAN to directly learn to perform instance segmentation.
The copy-paste GAN from~\cite{arandjelovic2019object} uses a generator to generate a mask that interpolates between two images and learns to discover objects. 
\cite{chen2019unsupervised} decompose the generative process in a segmentation step that separates foreground and background, and a generation step that in-paints the foreground segment.
Finally, \cite{bielski2019emergence} present a layered generator that distinguishes between foreground and background by perturbing the foreground image relative to the background.
Our approach is different insofar we present a structured generator that facilitates \emph{both} generation and segmentation of visual scenes that are composed of multiple objects and background, and which is trained purely for generation purposes.

\section{Experiments}
We investigate different aspects of the proposed structure on several multi-object datasets.
We are particularly interested in verifying that images are generated as compositions of objects and that the relational and background structure is properly utilized.
Moreover, we study how the incorporated structure affects the quality of the generated images, and how it may be used to perform inference in the form of instance segmentation.

\paragraph{Datasets}

We consider five multi-object datasets\footnote{Datasets are available online at \url{https://goo.gl/Eub81x}.}.
The first three are different variations of \emph{Multi-MNIST (MM)}, in which each image consists of three MNIST digits~\citep{lecun1998gradient} that were rescaled and drawn randomly onto a $64 \times 64$ canvas.
In \emph{Independent MM}, digits are chosen randomly and there is no relation among them.
The \emph{Triplet} variation imposes that all digits in an image are of the same type, requiring relations among the digits to be considered during the generative process.
Similarly in \emph{RGB Occluded MM} each image consists of exactly one red, green, and blue digit. 
The fourth dataset (\emph{CIFAR10 + MM}) is a variation of CIFAR10~\citep{krizhevsky2009learning} in which the digits from \emph{RGB Occluded MM} are drawn onto a randomly chosen (resized) CIFAR10 image.
Our final dataset is \emph{CLEVR}~\citep{johnson2017clevr}, which we downsample to $160 \times 240$ followed by center-cropping to obtain $128 \times 128$ images. 
Samples from each dataset can be seen in~\ref{app:samples}.

\paragraph{Evaluation}

A popular evaluation metric to evaluate GANs is the Fr\'{e}chet Inception Distance (FID)~\citep{heusel2017gans}.
It computes the distance between two empirical distributions of images as the Fr\'{e}chet distance between two corresponding multivariate Gaussian distributions that were estimated using the features of a pre-trained Inception network for each image.
Although prior work found that FID correlates well with perceived human quality of images on standard image datasets~\citep{lucic2018gans}, we find that FID is less useful when considering images consisting of \emph{multiple} salient objects.
Our results in Section~\ref{subsection:comparison} suggest that FID is not a good indicator of performance during later stages of training, and may easily be fooled by a GAN that focuses on image statistics rather than content (e.g. generating the correct number of objects).
We hypothesize that this inability is due to the Inception network having been trained only for single object classification.

Therefore, in addition to FID, we conduct two different studies among humans, 1) to compare images generated by our models to a baseline, and 2) to answer questions about the content of generated images.
The latter allows us to verify whether generated images are probable samples from the true image distribution.
As conducting a human evaluation of this kind is not feasible for large-scale hyper-parameter search, we will continue to rely on FID to select the ``best'' models during hyper-parameter selection.
Details of these human studies can be found in~\ref{app:experiment_details}.

\paragraph{Set-up}

\begin{wrapfigure}{r}{0.45\textwidth}
  \centering
  \begin{subfigure}{\textwidth}
  \includegraphics[width=0.45\textwidth]{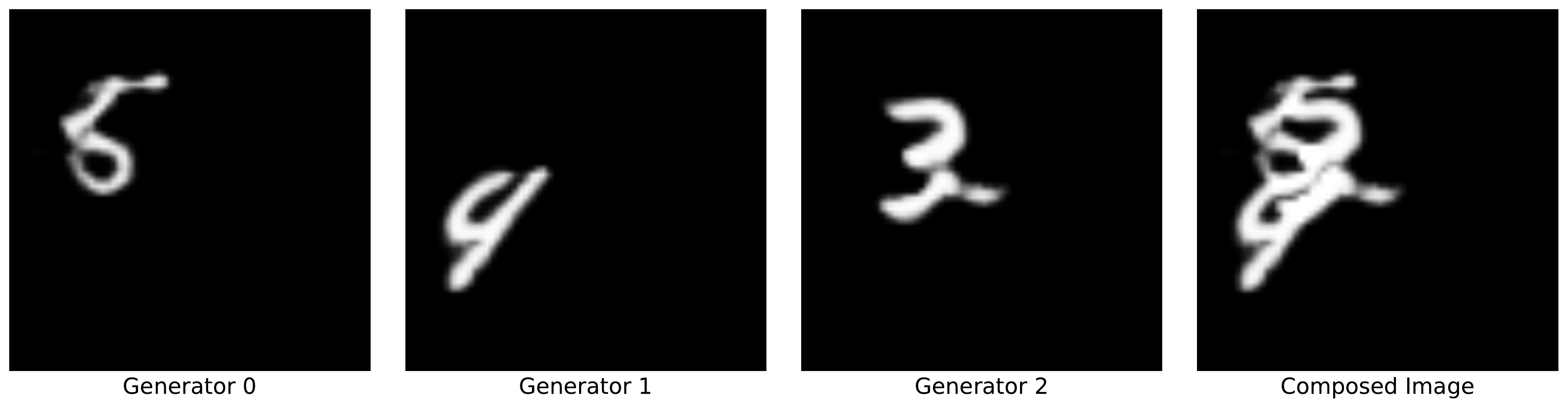}
  \end{subfigure}
  \begin{subfigure}{\textwidth}
  \includegraphics[width=0.45\textwidth]{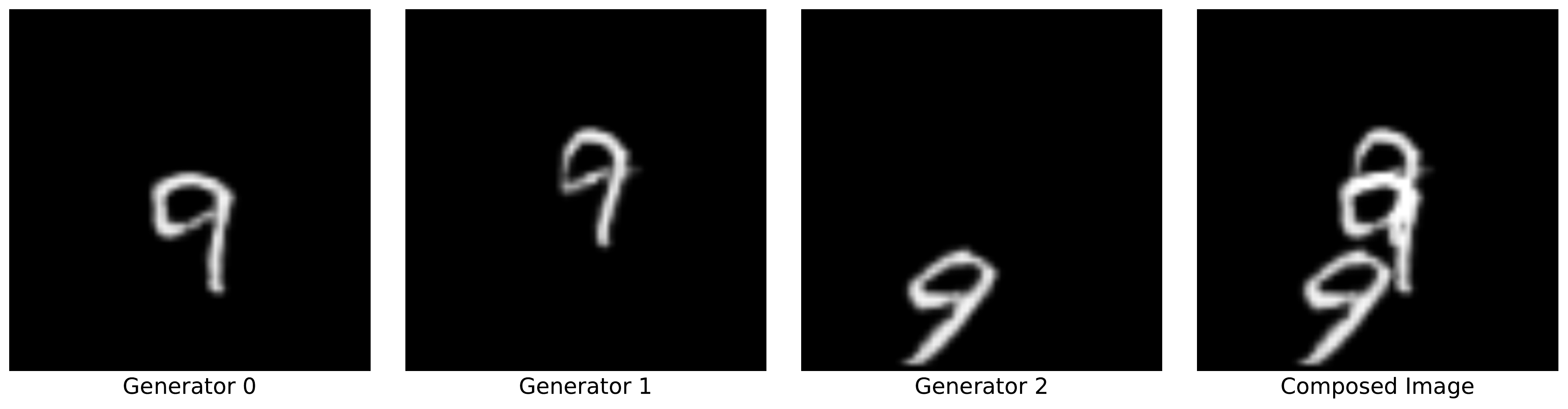}
  \end{subfigure}
  \begin{subfigure}{\textwidth}
  \includegraphics[width=0.45\textwidth]{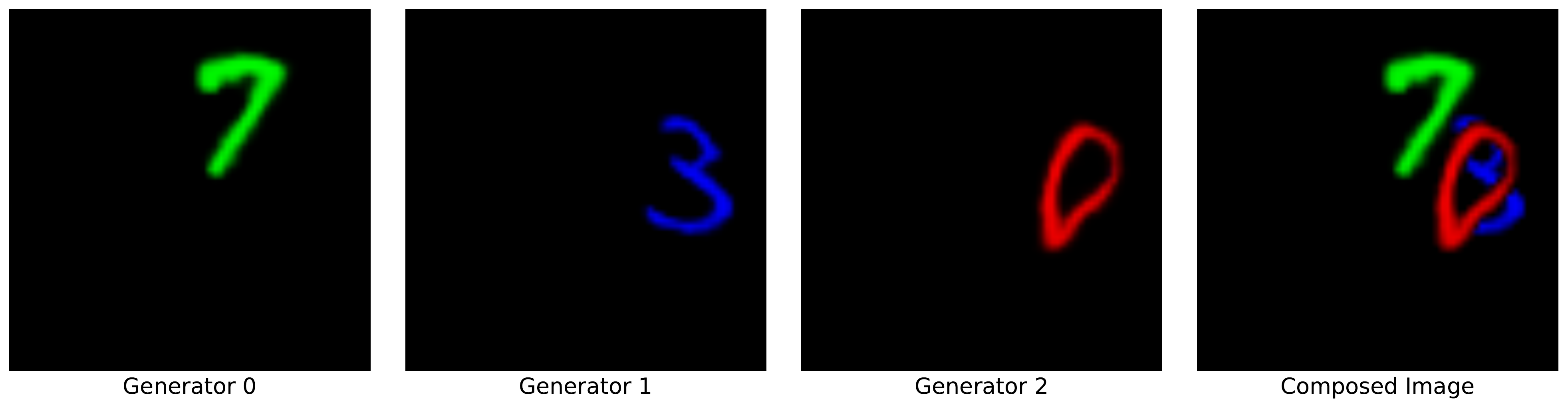}
  \end{subfigure}
  \caption{Generated images by \emph{3-GAN} on Multi-MNIST: \emph{Independent} (top), \emph{Triplet} (middle), and \emph{RGB Occluded} (bottom). The three columns on the left show the output of each object generator, and the right column the composed image.}
  \label{fig:mm-split}
  \vspace{-20pt}
\end{wrapfigure}

The GAN-based models are optimized with ADAM~\citep{kingma2015adam} using a learning rate of $10^{-4}$, and batch size $64$ for $1$M steps. 
We compute the FID (using $10$K samples) every $20$K steps, and select the best set of parameters accordingly.
On each dataset, we compare GANs that incorporate our proposed structure to a strong baseline that does not.
In both cases we conduct extensive grid searches covering on the order of 40-50 hyperparameter configurations for each dataset, using ranges that were previously found good for GANs~\citep{lucic2018gans,kurach2019large}.
Each configuration is ran with 5 different seeds to be able to estimate its variance.
A description of the hyper-parameter search and samples of our best models can be seen in~\ref{app:experiment_details} \&~\ref{app:samples}.
For \emph{IODINE}~\citep{greff2019multi}, we make use of the official trained model released by the authors.  

\paragraph{Notation}
In reporting our results we will break down the results obtained in terms of the structure that was incorporated in the generator.
We will denote \emph{k-GAN} to describe a generator consisting of $K=k$ components, \emph{k-GAN rel.} if it incorporates relational structure and \emph{k-GAN ind.} if it does not.
Additionally, we will append ``\emph{bg.}'' when the model includes a separate background generator.
We will use \emph{k-GAN} to refer more generally to GANs that incorporate any of the proposed structure, and \emph{GAN} to refer to the collection of GANs with different hyperparameters in our baseline.

\subsection{Qualitative Analysis}

\begin{figure}[t]
\centering
\begin{subfigure}[b]{\linewidth}
  \includegraphics[width=\linewidth]{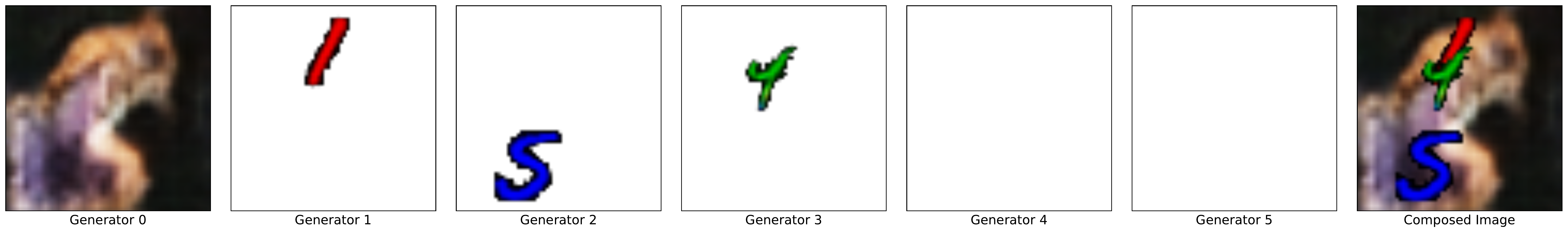}
\end{subfigure}
\begin{subfigure}[b]{\textwidth}
  \includegraphics[width=\linewidth]{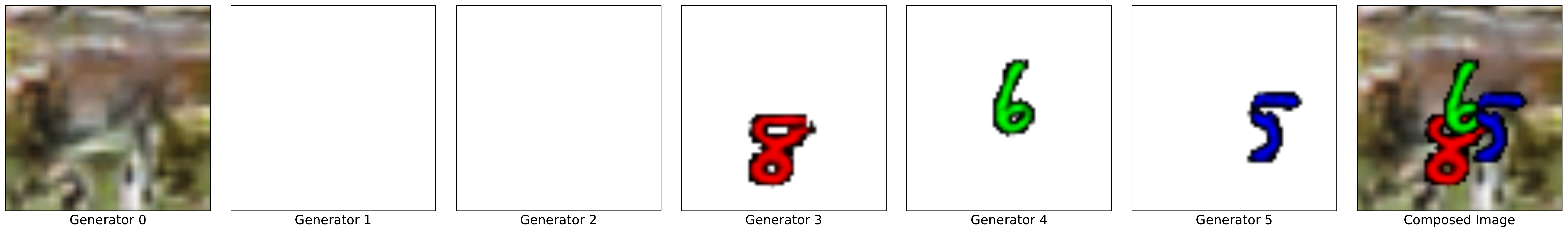}
\end{subfigure}
\begin{subfigure}[b]{\linewidth}
  \includegraphics[width=\linewidth]{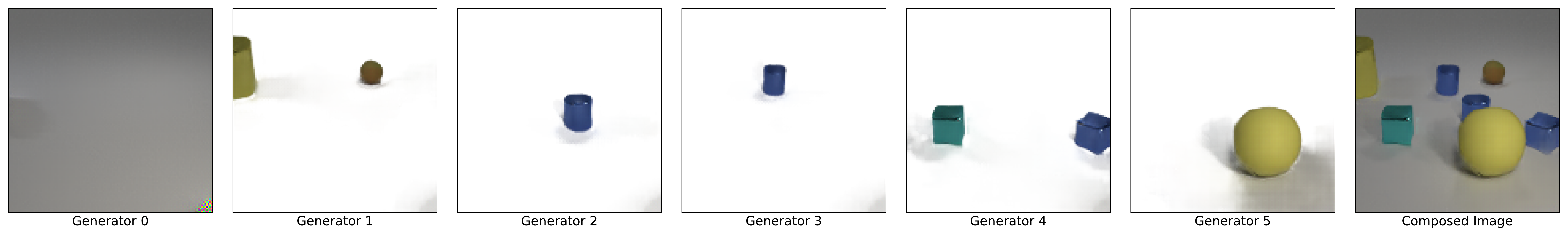}
\end{subfigure}
\begin{subfigure}[b]{\textwidth}
  \includegraphics[width=\linewidth]{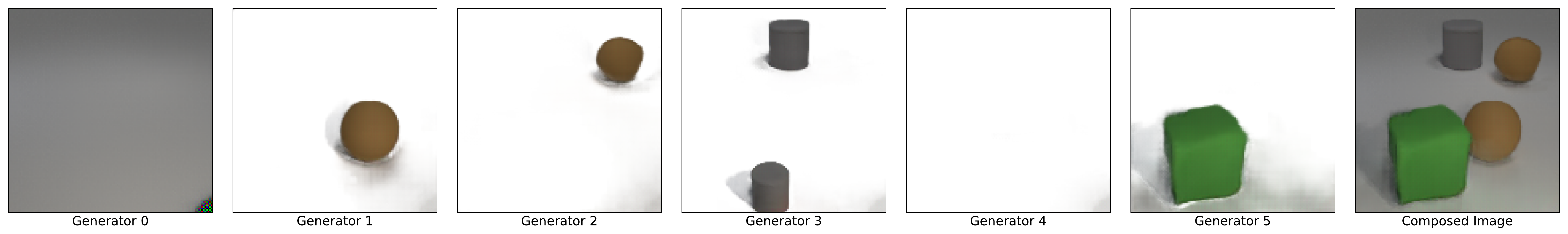}
\end{subfigure}
\caption{Generated samples by \emph{5-GAN rel. bg.} on \emph{CIFAR10 + MM} (top), and CLEVR (bottom). The left column corresponds to the output of the background generator. The next five columns are the outputs of each object generator, and the right column the composed image. Images are displayed as RGBA, with white denoting an alpha value of zero.}
\label{fig:relational_background_splits}
\end{figure}

\paragraph{Utilizing Structure}
We begin by analyzing the output of each (object) generator for \emph{k-GAN}.
Among the best performing models in our search, we consistently find that the final image is generated as a composition of images consisting of individual objects and background.
It can be seen that in the process of learning to generate images, \emph{k-GAN} learns about what are individual objects, and what is the background, without relying on prior knowledge or conditioning.
Using this learned knowledge suggests a natural approach to unsupervised instance segmentation, which we shall explore later.
Examples generated by \emph{k-GAN} for each dataset can be seen in Figure~\ref{fig:mm-split} and Figure~\ref{fig:relational_background_splits}.

On \emph{CLEVR}, where images often have a greater number of objects than the number of components $K$ that was used during training, we find that the generator continues to learn a factored solution (i.e. using visual primitives that consist of 1-3 objects).
This is interesting, as it suggests that using compositionality is preferable even when the factorization is sub-optimal.
A similar tendency was found when analyzing generated images by \emph{k-GAN ind.} when $k>3$ on Multi-MNIST.  
The generator decodes some latents as ``no digit'' in attempting to generate the correct number of digits.

From the generated samples by \emph{k-GAN rel.}, we observe that relations among the objects are correctly captured in most cases and that the relational mechanism can be used to generate the correct number of digits when $K$ is greater than the number of objects in the dataset.
We also observe that sometimes the background generator generates a \emph{single} object together with the background.
It rarely generates more than one object, which is further evidence that it is indeed more efficient to use the object generators.

\begin{figure}[b]
\centering
  \centering
  \includegraphics[width=\linewidth]{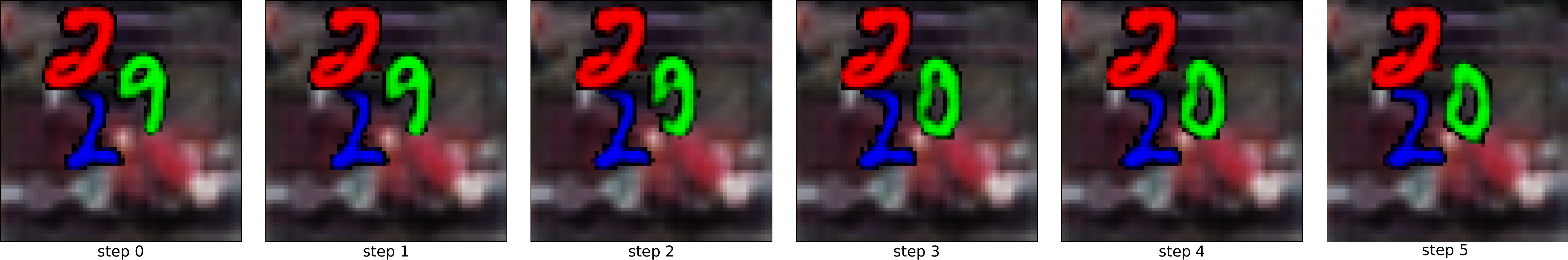}
  \centering
  \includegraphics[width=\linewidth]{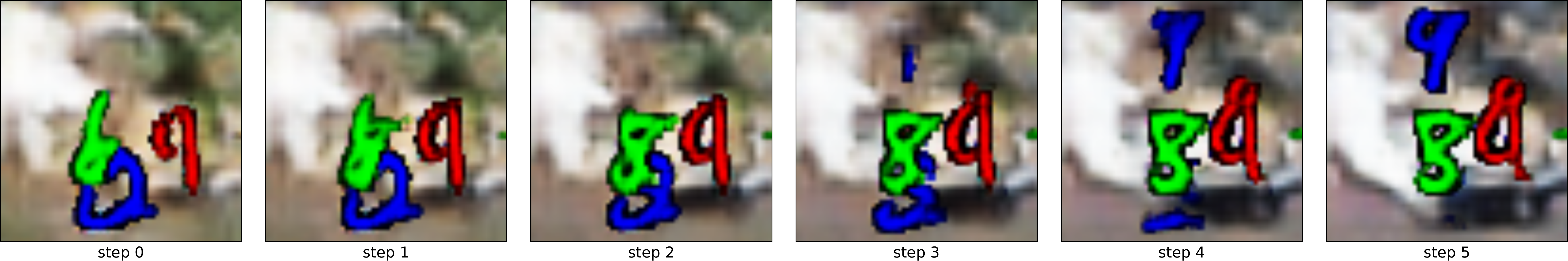}
\caption{Generated images by \emph{5-GAN rel. bg.} (top) and \emph{GAN} (bottom), when traversing the latent space of a single (object) generator. For \emph{k-GAN} only a single digit is transformed, while for \emph{GAN} the entire scene changes.}
\label{fig:relational}
\end{figure}

\paragraph{Latent Traversal}

We explore the degree to which the relational extension affects our initial independence assumption about objects.
If it were to cause the latent representations to become fully dependent on one another then it could negate the benefits of compositionality.
We conduct an experiment in which we traverse the latent space of a single latent vector in \emph{k-GAN rel.} by adding a random vector to the original sample with fixed increments, and generate images from the resulting latent vectors. 
An example can be seen in Figure~\ref{fig:relational} (top), where we find that traversing the latent space of a single component affects only the green digit, whereas the visual presentation of the others remains unaffected.

We observe this behavior (i.e. the relational mechanism does not unnecessarily interfere) for the majority of the generated samples, confirming to a large degree our own intuition of how the relational mechanism should be utilized. 
When traversing the latent space of \emph{GAN}, for which information about different objects is entangled, it results in a completely different scene (see Figure~\ref{fig:relational} bottom).
Hence, by disentangling the underlying representation it is more robust to common variations in image space.

\subsection{Quantitative Analysis}
\label{subsection:comparison}
\begin{figure}[t]
\centering
\includegraphics[width=\linewidth]{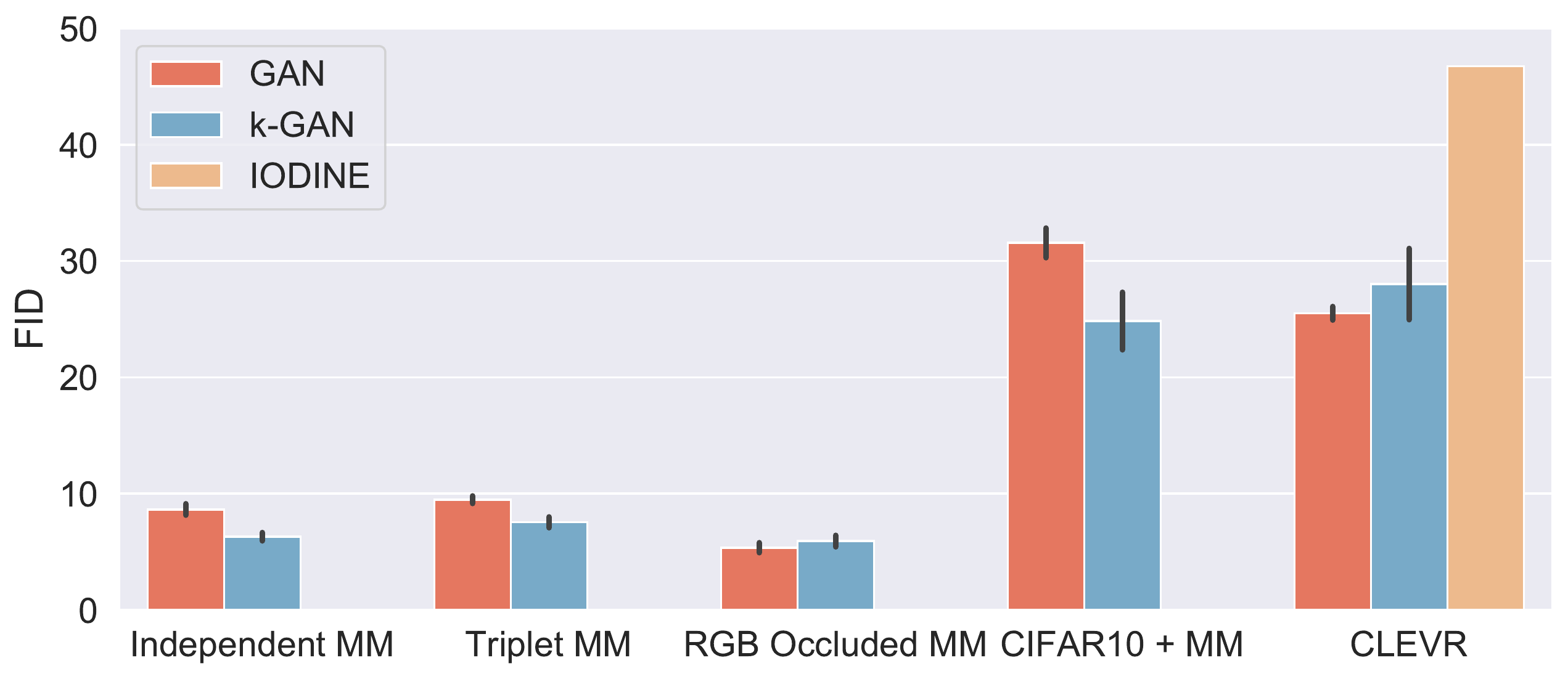}
\caption{The best FID obtained by \emph{GAN} and \emph{k-GAN} on all datasets following our grid search. The best configurations were chosen based on the smallest average FID (across 5 seeds). Standard deviations across seeds are illustrated with error bars. For \emph{IODINE} we made use of the official trained model for CLEVR released by the authors~\citep{greff2019multi}.}
\label{fig:fid_scores:all}
\end{figure}

\paragraph{Fr\'{e}chet Inception Distance (FID)} 
We train \emph{k-GAN} and \emph{GAN} on each dataset and compare the FID of the models with the lowest average FID across seeds (Figure~\ref{fig:fid_scores:all}).
On all datasets but \emph{CLEVR}, we find that \emph{k-GAN} compares better or similar to \emph{GAN}, although typically by a small margin.
Importantly, compared to the recent \emph{IODINE}~\citep{greff2019multi}, which also incorporates architectural structure to facilitate object compositionality, we observe that \emph{k-GAN} significantly outperforms.

An analysis, using different variations of \emph{k-GAN}, leads to several interesting observations.
On the relational datasets (Figure~\ref{fig:fid_scores:base-rel}) it can be observed that the relational extension is important to obtain good performance, while it does not harm performance when no relations are present. 
Likewise, on the background datasets (Figures~\ref{fig:fid_scores:bg} \&~\ref{fig:fid_scores:bg-no-bg}) we find that the background extension is important, although some mitigation is possible using the relational mechanism.
Finally, we observe several small differences in FID when changing the number of object generators $K$.
Surprisingly, we find that the lowest FID on \emph{Independent MM} is obtained by \emph{4-GAN} without relational structure, which by definition is unable to consistently generate 3 digits.
It suggests that FID is unable to capture these properties of images (motivating our human study), and renders any subtle FID differences between \emph{k-GAN} and \emph{GAN} inconclusive.

\begin{figure}[t]
\centering
\begin{subfigure}{.45\textwidth}
  \begin{subfigure}{0.45\textwidth}
  \centering
  \includegraphics[width=.85\linewidth]{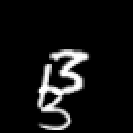}
\end{subfigure}%
\begin{subfigure}{0.45\textwidth}
  \centering
  \includegraphics[width=.85\linewidth]{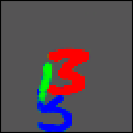}
\end{subfigure}
\caption{\emph{Independent MM}}
\end{subfigure}%
\begin{subfigure}{.45\textwidth}
  \begin{subfigure}{0.45\textwidth}
  \centering
  \includegraphics[width=.85\linewidth]{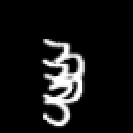}
\end{subfigure}%
\begin{subfigure}{0.45\textwidth}
  \centering
  \includegraphics[width=.85\linewidth]{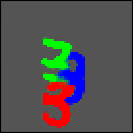}
\end{subfigure}
\caption{\emph{Triplet MM}}
\end{subfigure}
\caption{Result of segmenting unseen test-images using a segmenter that was trained on images (and labels) generated by \emph{3-GAN}. Note that this way of `inverting' the learned generative model is only possible due to the added structure, which makes the generative process interpretable and semantically understood.}
\label{fig:segmentation}
\end{figure}

\paragraph{Instance Segmentation}
We train a segmenter on data sampled from \emph{3-GAN} by treating the output of each object generator as pixel-level segmentation labels for the generated image (a similar technique was explored in \cite{spampinato2019adversarial} for motion segmentation in videos).
Note that this labeled data is obtained in a purely unsupervised fashion by exploiting the fact that, by incorporating structure, the learned generative process is now interpretable and semantically understood.
We test how the segmenter generalizes to real images (Figure~\ref{fig:segmentation}) for which we have ground-truth segmentations available and measure its accuracy using the Adjusted Rand Index (ARI)~\citep{hubert1985comparing} score.
We compare to training in purely supervised fashion on ground-truth data (further details are available in~\ref{app:experiment_details}).

In Table~\ref{table:segmentation} we find that using unsupervised data from \emph{3-GAN} is often as good as using ground-truth (and can even have a positive regularization effect).
These results strengthen our initial findings that \emph{k-GAN} reliably generates images as compositions of objects.
Moreover, it presents a novel approach to unsupervised instance segmentation that does not rely on an ``encoder'', but rather leverages the structure of the learned generative process to perform inference.

\paragraph{Human Evaluation}
We asked humans to compare the images generated by \emph{k-GAN rel.} to our \emph{GAN} baseline on \emph{RGB Occluded MM}, \emph{CIFAR10 + MM}, and \emph{CLEVR}, using the configuration with a background generator for the last two datasets\footnote{On \emph{CLEVR} we instructed the raters to ignore visual implausibilities due to floating objects (for both \emph{k-GAN} and \emph{GAN}) that may arise due to the fixed order in \eqref{eq:alpha_composit} and measured this effect separately in Figure~\ref{fig:human_eval_properties_clevr}.}.
For each model, we select the 10 best hyper-parameter configurations (lowest FID), from which we each generate 100 images.
We asked up to three raters for each image and report the majority vote or ``Equal'' if no decision was reached.

\begin{figure}[t]
\centering
\begin{subfigure}{0.5\linewidth}
  \centering
  \includegraphics[width=1\linewidth]{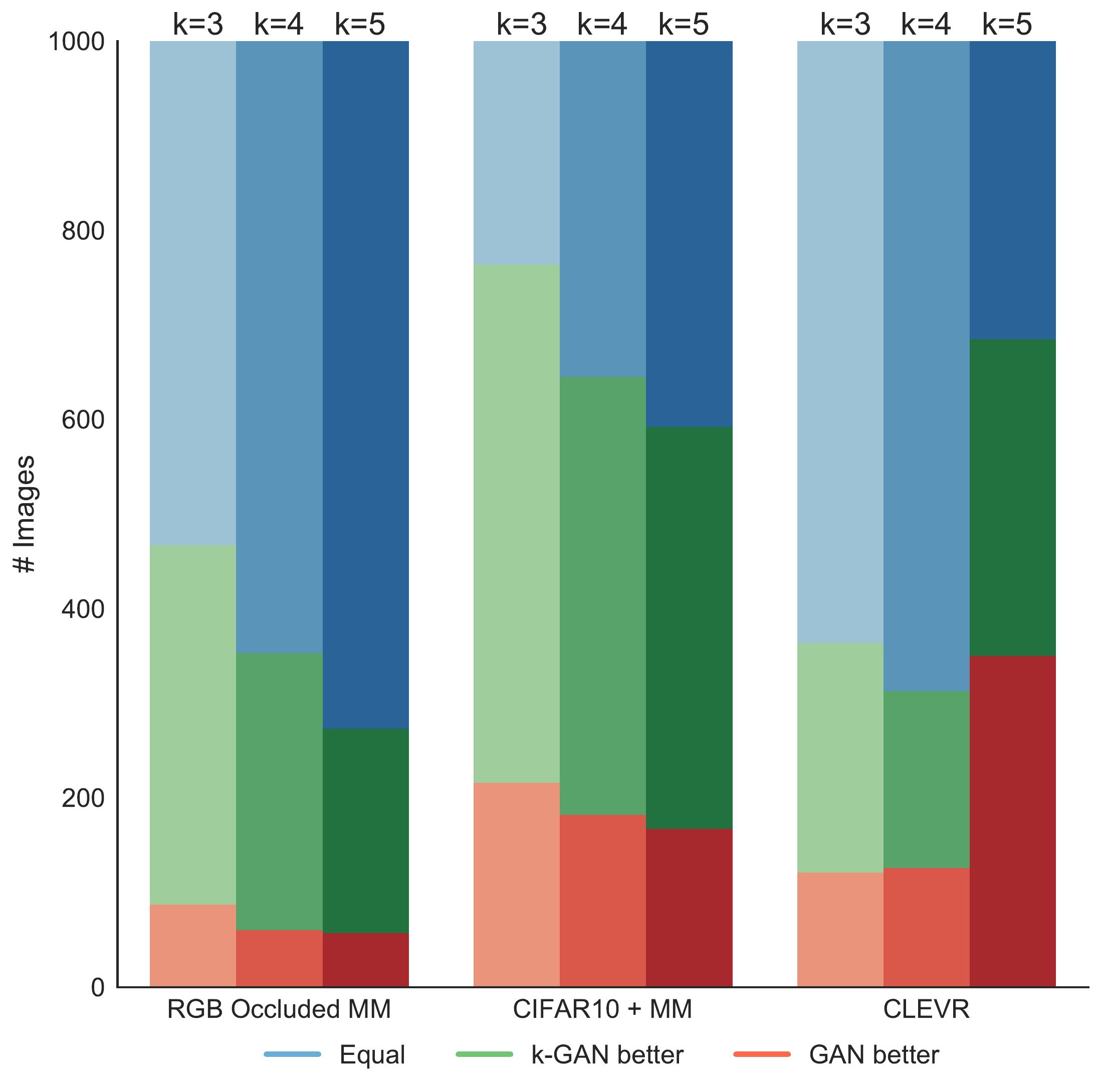}
  \caption{Comparing image quality}
  \label{fig:human_eval_subjective}
\end{subfigure}%
\begin{subfigure}{0.5\linewidth}
  \centering
  \includegraphics[width=1\linewidth]{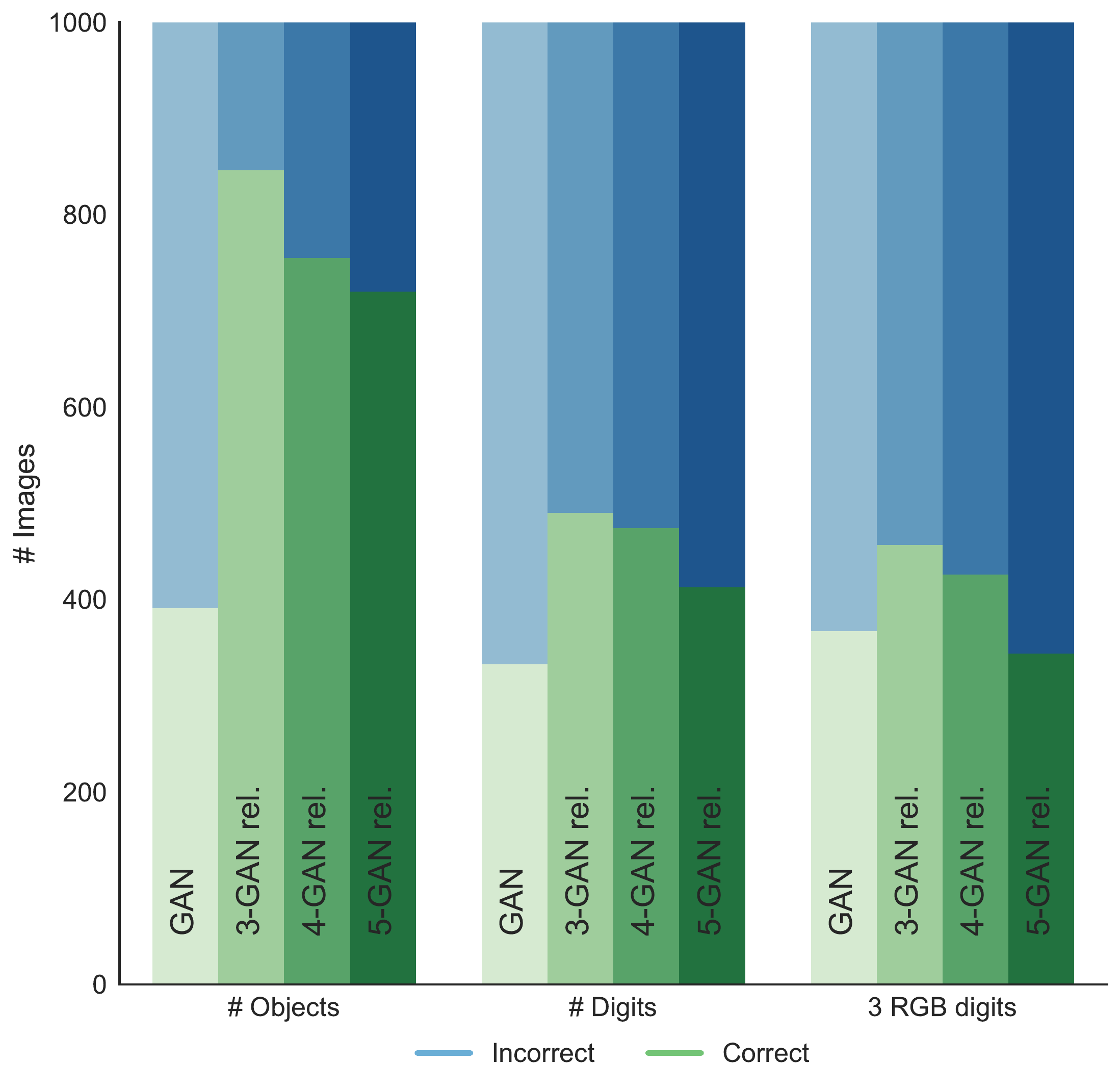}
  \caption{Properties on RGB Occluded MM}
  \label{fig:human_eval_properties_rgb_occluded}
\end{subfigure}
\caption{Results of human evaluation a) comparing the quality of the generated images by \emph{k-GAN} (k=3,4,5) to \emph{GAN} b) Properties of generated images by \emph{k-GAN} (k=3,4,5) and \emph{GAN} on \emph{RGB Occluded MM}. It can be seen that \emph{k-GAN} generates better images that are more faithful to the reference distribution.}
\end{figure}

Figure~\ref{fig:human_eval_subjective} reports the results when asking human raters to compare the visual quality of the generated images by \emph{k-GAN} to those by \emph{GAN}.
It can be seen that \emph{k-GAN} compares favorably across all datasets, and in particular on \emph{RGB Occluded MM} and \emph{CIFAR10 + MM} we observe large differences. 
We find that \emph{k-GAN} performs better even when $k>3$, which can be attributed to the relational mechanism, allowing all components to agree on the correct number of digits.

\begin{table}[t]
\centering
\begin{tabular}{lcc}
\toprule
      Dataset & \emph{Ground Truth} & \emph{3-GAN} \\
\midrule
      \emph{Independent MM} & 0.890 & 0.886 $\pm$ 0.003\\
      \emph{Triplet MM} & 0.911 & 0.903 $\pm$ 0.002 \\
      \emph{RGB Occluded MM} &  0.928 &   0.955 $\pm$ 0.003 \\
      \emph{CIFAR10 + MM} & 0.950 &  0.814  $\pm$ 0.131 \\
\end{tabular}
\caption{ARI scores obtained by training a segmenter on \emph{ground truth} data, and on samples from \emph{3-GAN}. Standard deviations are computed using generated data from the 5 best \emph{ 3-GAN} models according to FID.}
\label{table:segmentation}
\end{table}

In a second study we asked humans to report specific properties of the generated images, 
a complete list of which can be found in~\ref{app:experiment_details}.
Here our goal was to assess if the generated images by \emph{k-GAN} are more faithful to the reference distribution, which is particularly important in the context of representation learning.
The results on \emph{RGB Occluded MM} are summarized in Figure~\ref{fig:human_eval_properties_rgb_occluded}.
It can be seen that \emph{k-GAN} more frequently generates images that have the correct number of objects, number of digits, and that satisfy all properties simultaneously.
The difference between the correct number of digits and the correct number of objects suggests that the generated objects are often not recognizable as digits.
This does not appear to be the case from the generated samples in \ref{app:samples}, suggesting that the raters may not have been familiar enough with the variety of MNIST digits.

On \emph{CIFAR10 + MM} (Figure~\ref{fig:human_eval_properties_cifar10}) it appears that \emph{GAN} is able to accurately generate the correct number of objects, although the addition of background makes it difficult to provide a comparison in this case.
Indeed, on the number of digits, \emph{k-GAN} outperforms \emph{GAN} by the same margin as one would expect compared to the results in Figure~\ref{fig:human_eval_properties_rgb_occluded}.

Finally, in comparing the generated images by \emph{k-GAN} and \emph{GAN} on \emph{CLEVR}, we noticed that the former generated more crowded scenes (containing multiple large objects in the center), and more frequently generated objects with distorted shapes or mixed colors. 
On the other hand, we found cases in which \emph{k-GAN} generated scenes containing ``flying'' objects, a by-product of the fixed order in which we apply \eqref{eq:alpha_composit}.
We asked humans to score images based on these properties, which confirmed these observations (see Figure~\ref{fig:human_eval_properties_clevr}).

\section{Discussion}
The experimental results confirm that the proposed structure is beneficial in generating images of multiple objects, and is utilized according to our own intuitions.
By structuring the generative process, the generator learns about objects (and background) without requiring any supervision. 
Notably, we were able to extract this knowledge by `inverting' the generative model to learn how to perform instance segmentation (and from which it is easy to obtain corresponding object representations).
Indeed, although the computational complexity of our approach scales linearly with $K$ (while the number of parameters stays constant due to sharing), these provide concrete reasons for why one would want to pursue this direction compared to other neural approaches to image generation and representation learning that do not incorporate structure. 

In general, we find that it is encouraging to see that a simple inductive bias applied to standard neural network generator yields state of the art image generation capabilities compared to other purely unsupervised object-centric approaches. 
Especially with similar approaches requiring a lot more engineering to achieve worse results~\citep{eslami2016attend,greff2017neural,greff2019multi}.  
In that sense, it is likely that even better results can be obtained by building on the foundation as is presented here.
Conversely, we also expect that some of the innovations used here, such as a separate background generator, a relational mechanism, and compositing can be used to improve other approaches.

In order to benefit maximally from the presented structure, it is desirable to be able to accurately estimate the (minimum) number of objects in the images in advance.
This task is ill-posed as it relies on a precise definition of ``object'' that is generally not available.
In our experiments on \emph{CLEVR} we encountered a similar situation in which the number of components used for training was often smaller than the number of objects in the images.
Interestingly, we found that it does not render the incorporated structure meaningless, but that the generator discovers ``primitives'' that correspond to multiple objects as in Figure~\ref{fig:relational_background_splits}.

One area of improvement is in being able to accurately determine foreground, and background when combining the outputs of the object generators using alpha compositing.
On \emph{CLEVR} we observed cases in which objects appear to be flying, which is the result of being unable to route the information content of a ``foreground'' object to the corresponding ``foreground'' generator as induced by the fixed order in which images are composed.
Although in principle the relational mechanism may account for this distinction, it may be easier to learn this when a more explicit mechanism is incorporated.

Another interesting avenue for improvement is to also incorporate structure in the discriminator, which acts as a loss function in the context of GANs.
As we found that the pre-trained \emph{Inception} embedding is limited in reasoning about the validity of multi-object images, the discriminator may experience similar difficulties in accurately judging images from being real or fake without additional structure.
Ideally, we would have the discriminator reason about the correctness of each object individually, as well as the image as a whole.
Adding additional `patch discriminators'~\citep{isola2017image}, where patches correspond to objects may serve a starting point in pursuing this direction.

\section{Conclusion}
We have investigated the usefulness (and the feasibility) of compositionality at the representational level of objects in GANs through a simple modification of a standard generator.
The resulting \emph{structured} GAN can be trained in a purely unsupervised manner as it does not rely on auxiliary inputs, and on a benchmark of multi-object datasets we have shown that our generative model learns about individual objects in the process of synthesizing samples.
We have also shown how relations between objects, and background can be modeled by incorporating two modular extensions. 
A human study revealed that this leads to a better generative model of images compared to a strong baseline of GANs.
Additionally, on the challenging CLEVR dataset~\citep{johnson2017clevr}, it was shown how our approach is able to improve over IODINE~\citep{greff2019multi} in terms of image generation capabilities, and a similar observation can be made for earlier purely unsupervised object-centric approaches, e.g. N-EM~\citep{greff2017neural} or AIR~\citep{eslami2016attend} by virtue of their inherent limitations.

The ultimate goal of this work is to learn structured generative models that can be inverted to perform representation learning.
To this extent, we have proposed a novel approach to leverage the structure of the generator to generate data that can then be used to train an instance segmenter in a supervised fashion.
We have shown that this segmenter generalizes to real images, which allows one to acquire corresponding object representations.
By using GANs, we were able to overcome the limitations of other recent purely unsupervised object-centric approaches that do not scale to more complex multi-object images.

\section*{Acknowledgements}

The authors wish to thank Damien Vincent, Alexander Kolesnikov, Olivier Bachem, Klaus Greff, and Paulo Rauber for helpful comments and
constructive feedback.
The authors are grateful to Marcin Michalski and Pierre Ruyssen for their technical support.
This research was in part supported by the Swiss National Science Foundation grant 200021\_165675/1, and by hardware donations from NVIDIA Corporation as part of the Pioneers of AI Research award, and by IBM.

\bibliographystyle{iclr2019_conference}
\bibliography{references}

\clearpage
\appendix                                     

\section{Additional Experiment Results}
\label{app:experiment_results}
\subsection{FID Study}

\begin{figure}[h]
\centering
\includegraphics[width=\linewidth]{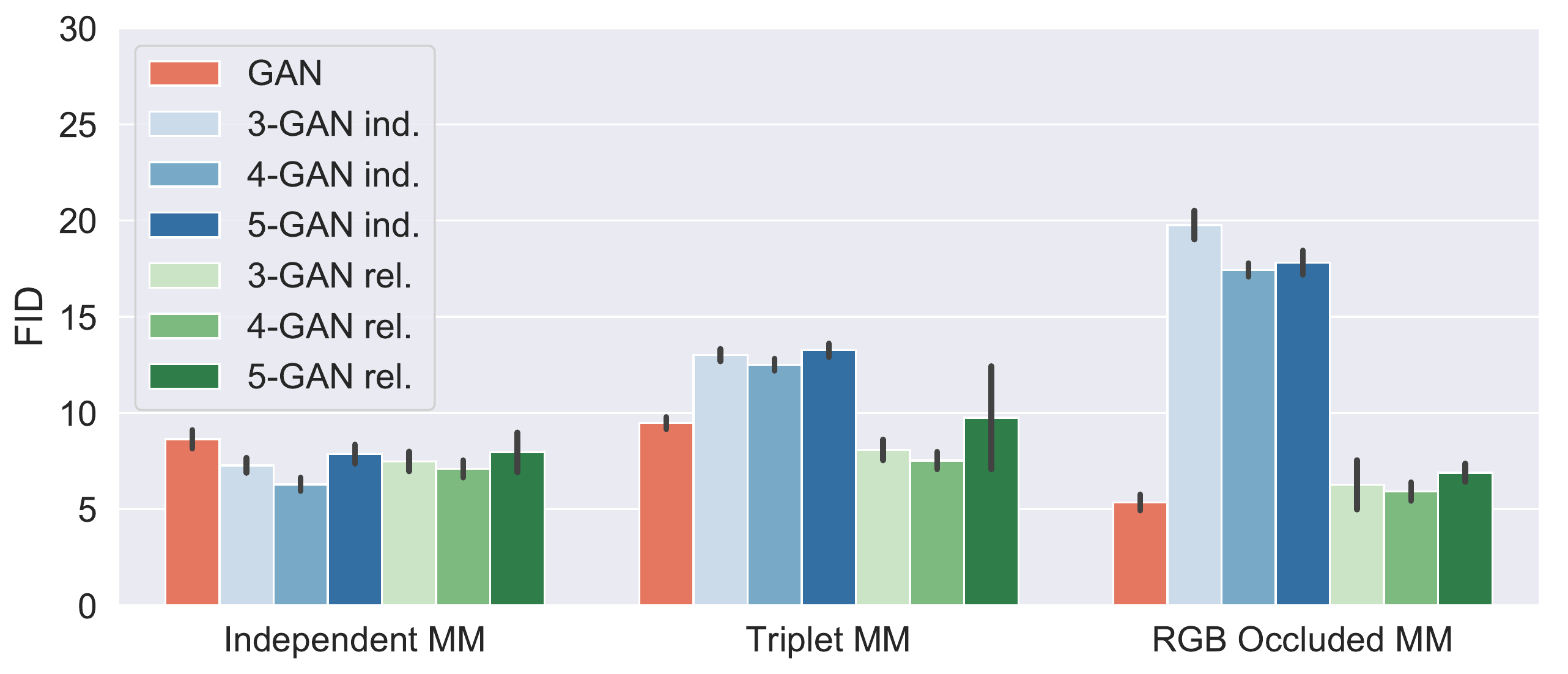}
\caption{\textbf{Analysis.} The best FID obtained by \emph{GAN} and \emph{k-GAN} on \emph{Independent MM}, \emph{Triplet MM}, and \emph{RGB Occluded MM} following our grid search. The best configurations were chosen based on the smallest average FID (across 5 seeds). Standard deviations across seeds are illustrated with error bars.}
\label{fig:fid_scores:base-rel}
\end{figure}

\begin{figure}[h]
\centering
\includegraphics[width=0.75\linewidth]{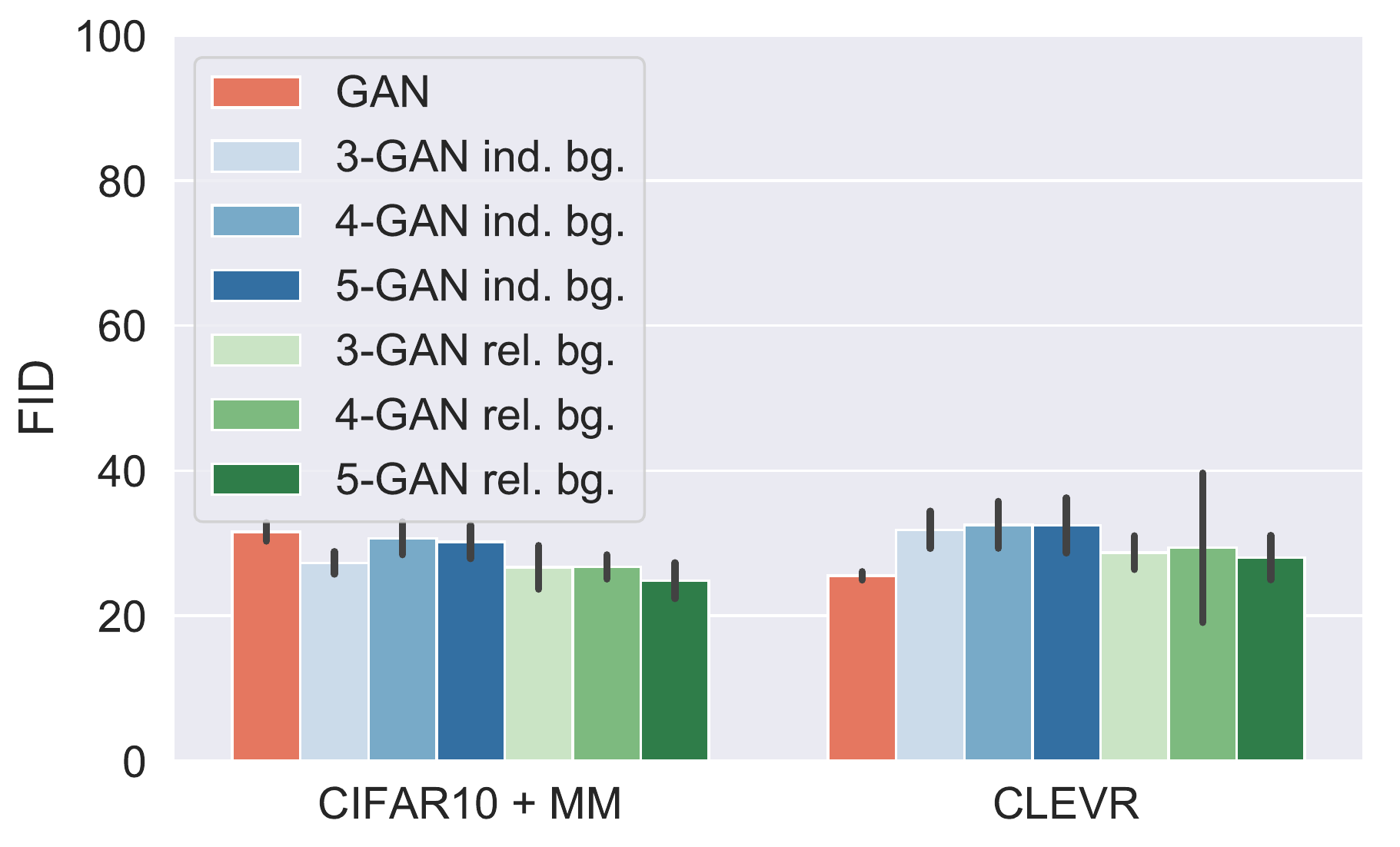}
\caption{\textbf{Analysis.} The best FID obtained by \emph{GAN} and \emph{k-GAN} on \emph{MM + CIFAR10}, and \emph{CLEVR} following our grid search. In this figure all \emph{k-GAN} variations make use of the background extension. The best configurations were chosen based on the smallest average FID (across 5 seeds). Standard deviations across seeds are illustrated with error bars.}
\label{fig:fid_scores:bg}
\end{figure}

\begin{figure}[h]
\centering
\includegraphics[width=0.75\linewidth]{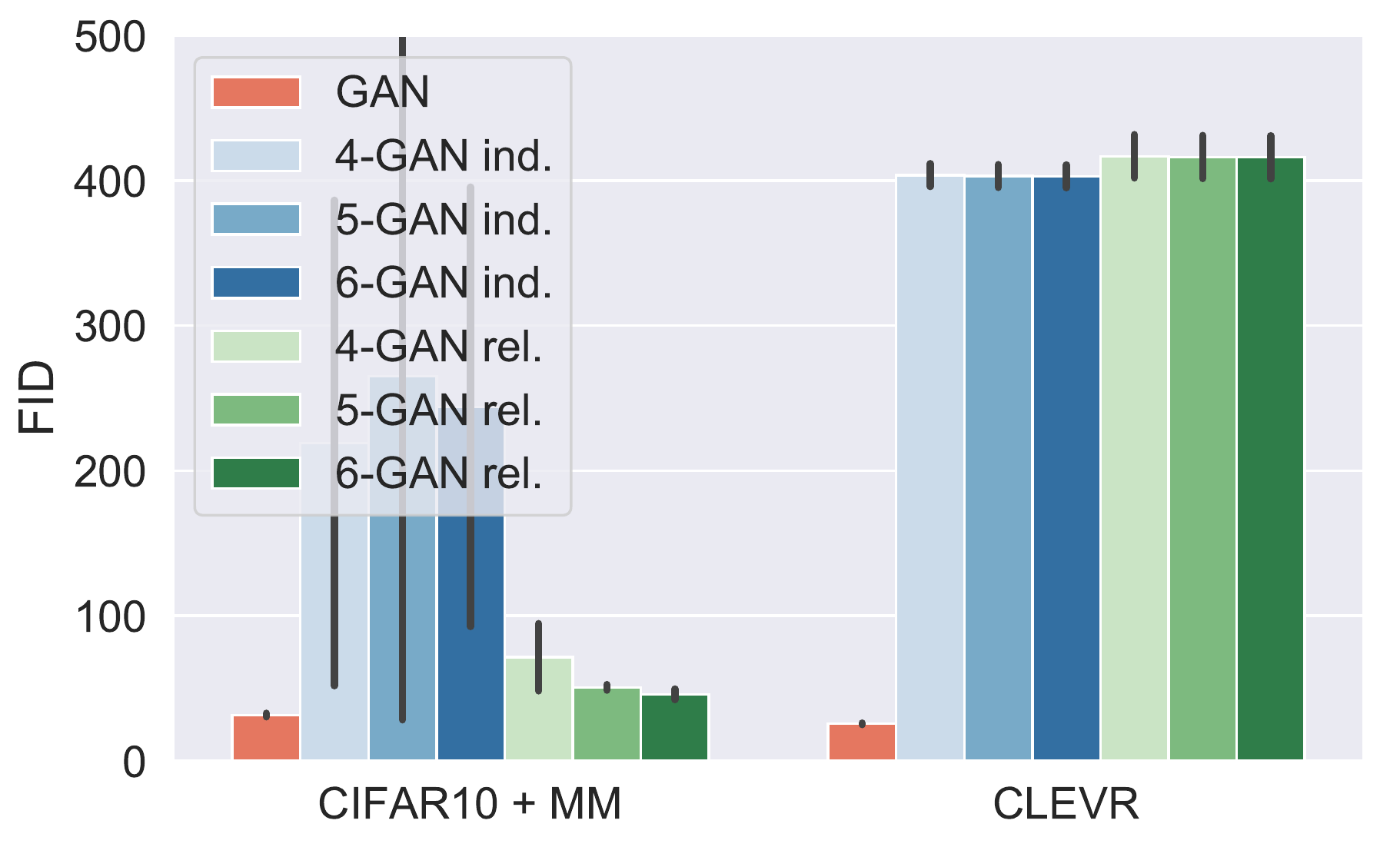}
\caption{\textbf{Analysis.} The best FID obtained by \emph{GAN} and \emph{k-GAN} on \emph{MM + CIFAR10}, and \emph{CLEVR} following our grid search. In this figure all \emph{k-GAN} variations do not use the background extension. The best configurations were chosen based on the smallest average FID (across 5 seeds). Standard deviations across seeds are illustrated with error bars.}
\label{fig:fid_scores:bg-no-bg}
\end{figure}

Figures~\ref{fig:fid_scores:base-rel}--\ref{fig:fid_scores:bg-no-bg} and Table~\ref{table:best_models}.

\begin{table}[h]
\centering
\begin{tabular}{llllllllrrr}
\toprule
      model & gan type & norm. & penalty & blocks & heads &  share &  bg. int. &  $\beta_1$ &  $\beta_2$ &  $\lambda$ \\
\midrule
        GAN &   NS-GAN &    spec. &    none &      x &     x &  x &       x &    0.5 &  0.999 &      10 \\
 3-GAN ind. &   NS-GAN &    spec. &    WGAN &      x &     x &  x &       x &    0.9 &  0.999 &       1 \\
 4-GAN ind. &   NS-GAN &    spec. &    WGAN &      x &     x &  x &       x &    0.9 &  0.999 &       1 \\
 5-GAN ind. &   NS-GAN &    spec. &    WGAN &      x &     x &  x &       x &    0.9 &  0.999 &       1 \\
 3-GAN rel. &   NS-GAN &    spec. &    WGAN &      1 &     1 &  no &       x &    0.9 &  0.999 &       1 \\
 4-GAN rel. &   NS-GAN &    spec. &    WGAN &      1 &     1 &  no &       x &    0.9 &  0.999 &       1 \\
 5-GAN rel. &   NS-GAN &    spec. &    WGAN &      2 &     1 &  no &       x &    0.9 &  0.999 &       1 \\
\midrule
        GAN &   NS-GAN &    spec. &    none &      x &     x &  x &       x &    0.5 &  0.999 &       1 \\
 3-GAN ind. &   NS-GAN &    spec. &    WGAN &      x &     x &  x &       x &    0.9 &  0.999 &       1 \\
 4-GAN ind. &   NS-GAN &    spec. &    WGAN &      x &     x &  x &       x &    0.9 &  0.999 &       1 \\
 5-GAN ind. &   NS-GAN &    spec. &    WGAN &      x &     x &  x &       x &    0.9 &  0.999 &       1 \\
 3-GAN rel. &   NS-GAN &    spec. &    WGAN &      1 &     1 &  no &       x &    0.9 &  0.999 &       1 \\
 4-GAN rel. &   NS-GAN &    spec. &    WGAN &      1 &     2 &  no &       x &    0.9 &  0.999 &       1 \\
 5-GAN rel. &   NS-GAN &    spec. &    WGAN &      2 &     1 &  no &       x &    0.9 &  0.999 &       1 \\
\midrule
        GAN &   NS-GAN &    spec. &    none &      x &     x &  x &       x &    0.5 &  0.999 &       1 \\
 3-GAN ind. &   NS-GAN &    spec. &    WGAN &      x &     x &  x &       x &    0.9 &  0.999 &       1 \\
 4-GAN ind. &   NS-GAN &    spec. &    WGAN &      x &     x &  x &       x &    0.9 &  0.999 &       1 \\
 5-GAN ind. &   NS-GAN &    spec. &    WGAN &      x &     x &  x &       x &    0.9 &  0.999 &       1 \\
 3-GAN rel. &   NS-GAN &        none &    WGAN &      2 &     2 &   yes &       x &    0.9 &  0.999 &       1 \\
 4-GAN rel. &   NS-GAN &        none &    WGAN &      2 &     2 &  no &       x &    0.9 &  0.999 &       1 \\
 5-GAN rel. &   NS-GAN &        none &    WGAN &      2 &     2 &   yes &       x &    0.9 &  0.999 &       1 \\
\midrule
            GAN &   NS-GAN &        none &    WGAN &      x &     x &  x &       x &    0.9 &  0.999 &       1 \\
 3-GAN ind. bg. &   NS-GAN &        none &    WGAN &      x &     x &  x &        x &    0.9 &  0.999 &       1 \\
 4-GAN ind. bg. &   NS-GAN &        none &    WGAN &      x &     x &  x &        x &    0.9 &  0.999 &       1 \\
 5-GAN ind. bg. &   NS-GAN &        none &    WGAN &      x &     x &  x &       x &    0.9 &  0.999 &       1 \\
 3-GAN rel. bg. &   NS-GAN &        none &    WGAN &      2 &     1 &   yes &        yes &    0.9 &  0.999 &       1 \\
 4-GAN rel. bg. &   NS-GAN &        none &    WGAN &      2 &     1 &   yes &        yes &    0.9 &  0.999 &       1 \\
 5-GAN rel. bg. &   NS-GAN &        none &    WGAN &      2 &     2 &   yes &       no &    0.9 &  0.999 &       1 \\
\midrule
            GAN &     WGAN &        none &    WGAN &      x &     x &  x &       x &    0.9 &  0.999 &       1 \\
 3-GAN ind. bg. &   NS-GAN &        none &    WGAN &      x &     x &  x &       x &    0.9 &  0.999 &       1 \\
 4-GAN ind. bg. &   NS-GAN &        none &    WGAN &      x &     x &  x &        x &    0.9 &  0.999 &       1 \\
 5-GAN ind. bg. &   NS-GAN &        none &    WGAN &      x &     x &  x &       x &    0.9 &  0.999 &       1 \\
 3-GAN rel. bg. &   NS-GAN &        none &    WGAN &      2 &     1 &  no &        yes &    0.9 &  0.999 &       1 \\
 4-GAN rel. bg. &   NS-GAN &        none &    WGAN &      1 &     2 &  no &       no &    0.9 &  0.999 &       1 \\
 5-GAN rel. bg. &   NS-GAN &        none &    WGAN &      2 &     2 &  no &       no &    0.9 &  0.999 &       1 \\
\bottomrule
\end{tabular}
\caption{best hyper-parameter configuration for each model that were obtained following our grid search.
Configurations were chosen based on the smallest average FID (across 5 seeds) as reported in Figure~\ref{fig:fid_scores:all}.
Each block corresponds to a dataset (from top to bottom: \emph{Independent MM}, \emph{Triplet MM}, \emph{RGB Occluded MM}, \emph{CIFAR10 + MM}, \emph{CLEVR}).}
\label{table:best_models}
\end{table}

\clearpage
\subsection{Human Study Properties}

\begin{figure}[h]
\centering
\includegraphics[width=0.8\linewidth]{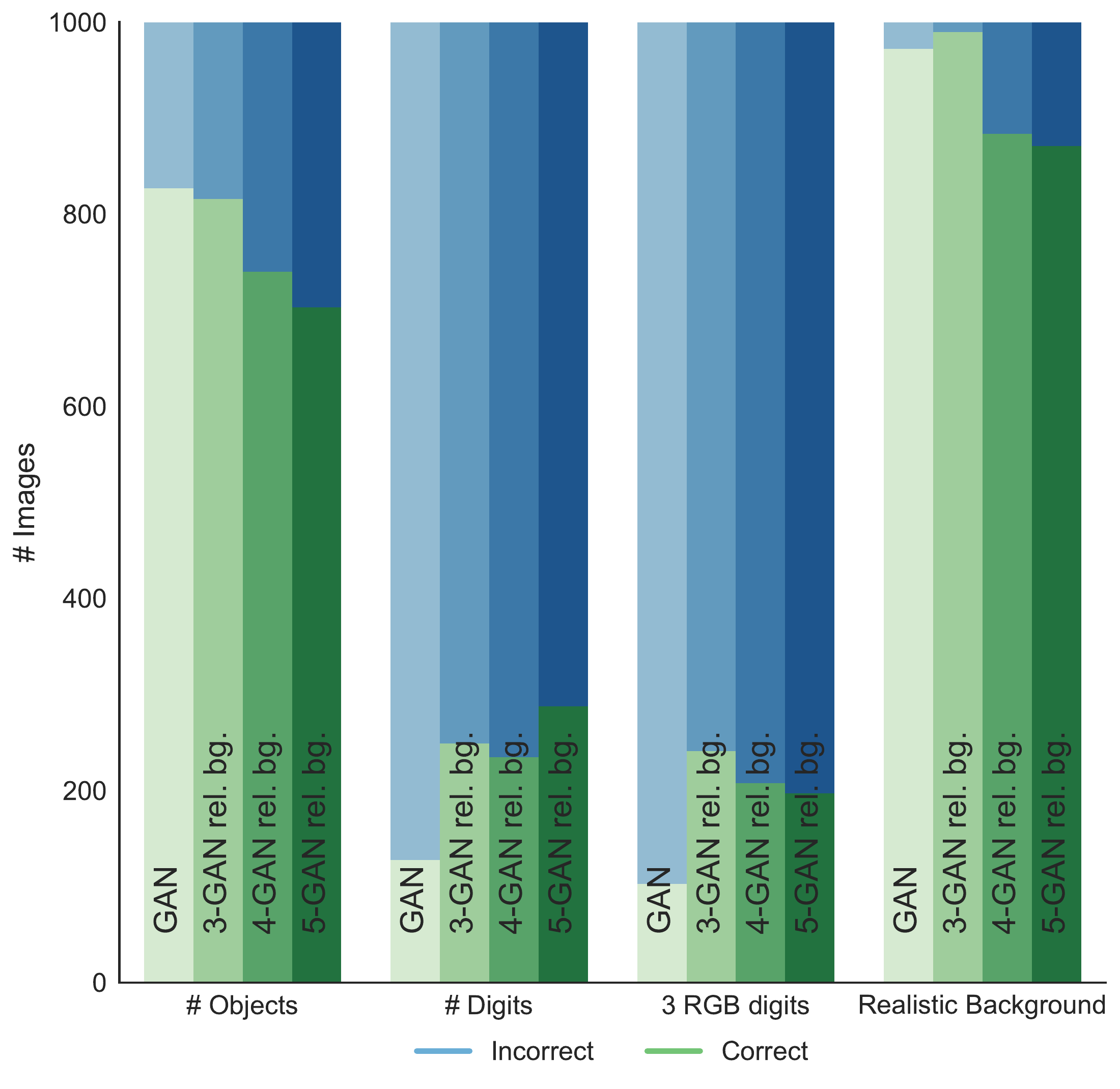}
\caption{Additional results of human evaluation. Properties of generated images by \emph{k-GAN} (k=3,4,5) and \emph{GAN} on \emph{CIFAR10 + MM}.}
\label{fig:human_eval_properties_cifar10}
\end{figure}

\begin{figure}[h]
\centering
\includegraphics[width=0.8\linewidth]{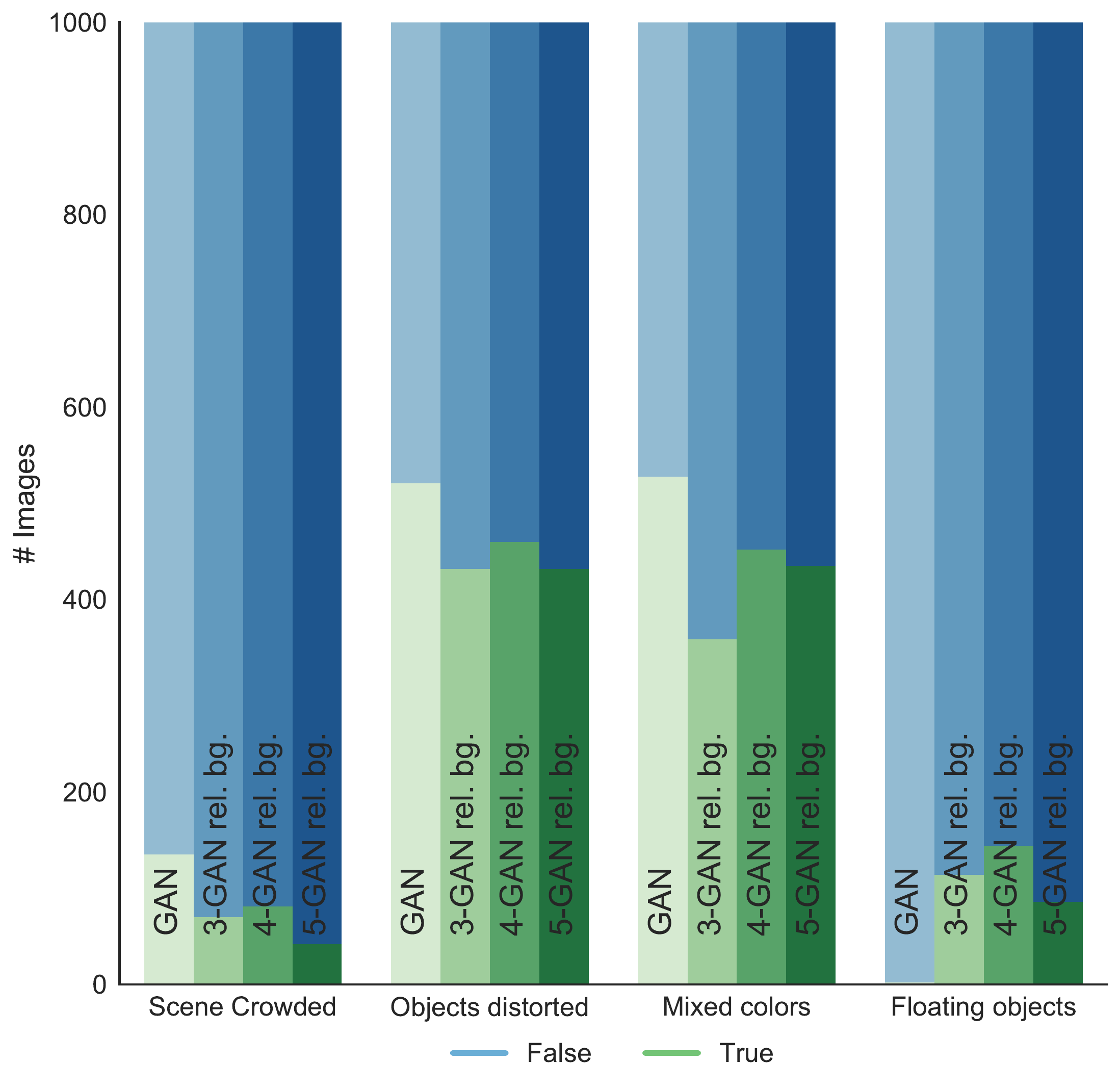}
\caption{Additional results of human evaluation. Properties of generated images by \emph{k-GAN} (k=3,4,5) and \emph{GAN} on CLEVR. Note that on CLEVR all evaluated properties are undesirable, and thus a larger number of ``False'' responses is better.}
\label{fig:human_eval_properties_clevr}
\end{figure}

\begin{figure}[h]
\centering
\includegraphics[width=\linewidth]{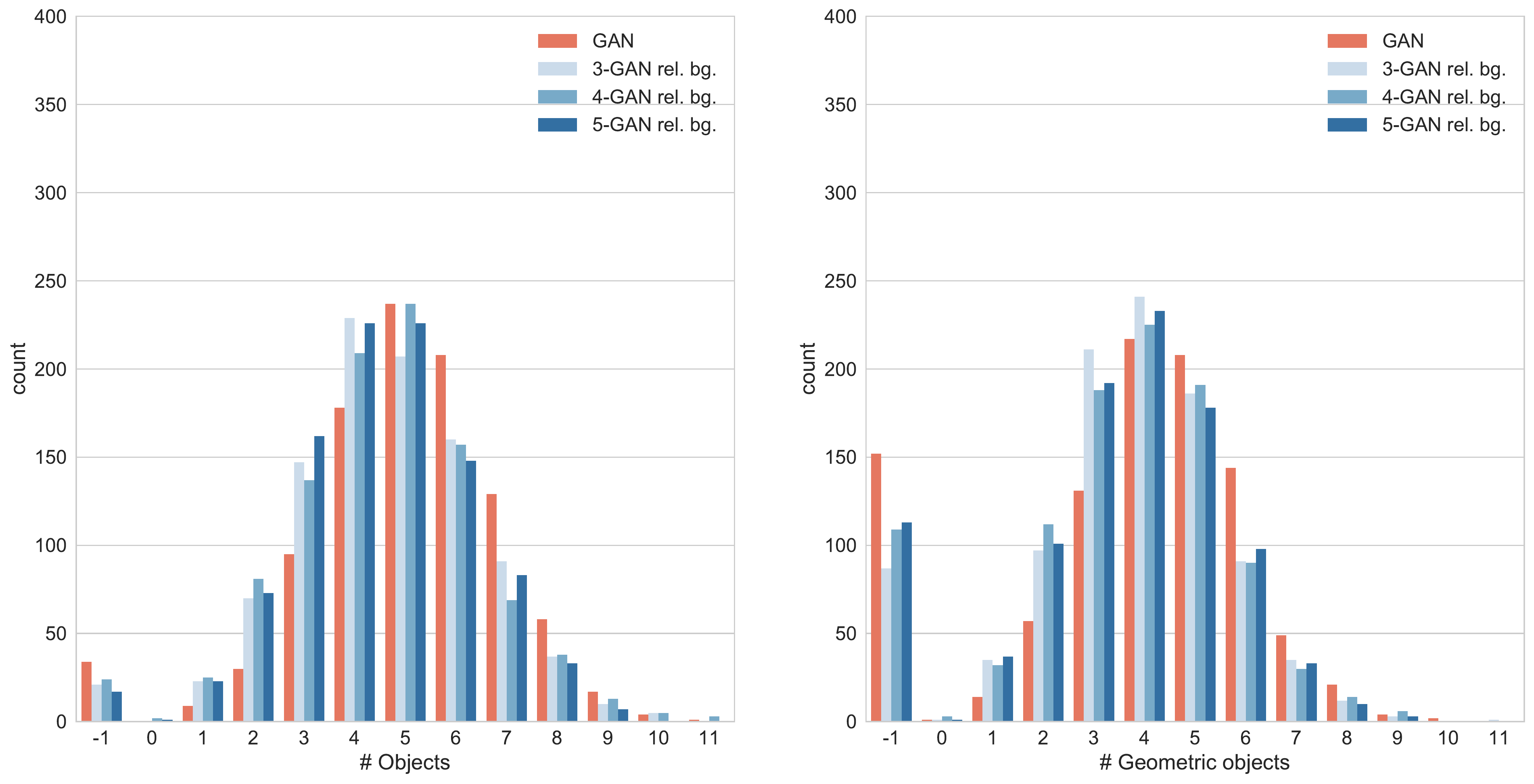}
\caption{Results of human evaluation. Number of (geometric) objects in generated images by \emph{k-GAN} (k=3,4,5) and \emph{GAN} on CLEVR. A value of -1 implies a majority vote could not be reached.}
\label{fig:human_eval_properties_clevr_counts}
\end{figure}

Figures~\ref{fig:human_eval_properties_cifar10}--\ref{fig:human_eval_properties_clevr_counts}.

\clearpage
\section{Experiment Details}
\label{app:experiment_details}
\subsection{Model specifications}
The generator and discriminator neural network architectures in all our experiments are based on DCGAN~\citep{radford2015unsupervised}.

\paragraph{Object Generators} 
\emph{k-GAN ind.} introduces $K=k$ copies of an object generator (i.e. tied weights, DCGAN architecture) that each generate an image from an independent sample of a 64-dimensional \emph{UNIFORM(-1, 1)} prior $P(Z)$.

\paragraph{Relational Structure}
When a relational stage is incorporated (\emph{k-GAN rel.}) each of the $\bm{z}_i \sim P(Z)$ is first updated, before being passed to the generators. 
These updates are computed using one or more \emph{attention blocks}, which integrate Multi-Head Dot-Product Attention (MHDPA)~\citep{vaswani2017attention} with a post-processing step~\citep{zambaldi2018deep}.
A single head of an attention block updates $\bm{z}_i$ according to \eqref{eq:attention_block-1},~\eqref{eq:attention_block-2}, and~\eqref{eq:attention_block-3}.

In our experiments, we use a single-layer neural network (fully-connected, 32 ReLU) followed by LayerNorm~\citep{ba2016layer} for each of $\text{MLP}^{\textit{(q)}}$, $\text{MLP}^{\textit{(k)}}$, $\text{MLP}^{\textit{(v)}}$.
We implement $\text{MLP}^{\textit{(u)}}$ with a two-layer neural network (each fully-connected, 64 ReLU), and apply LayerNorm after summing with $\bm{z}_i$.
Different heads in the same block use different parameters for $\text{MLP}^{\textit{(q)}}$, $\text{MLP}^{\textit{(k)}}$, $\text{MLP}^{\textit{(v)}}$, $\text{MLP}^{\textit{(u)}}$.
If multiple heads are present, then their outputs are concatenated and transformed by a single-layer neural network (fully-connected, 64 ReLU) followed by LayerNorm to obtain the new $\bm{\hat{z}}_i$.
If the relational stage incorporates multiple attention blocks that iteratively update $\bm{z}_i$, then we consider two variations: using unique weights for each MLP in each block or sharing their weights across blocks.

\paragraph{Background Generation}
When a background generator is incorporated (eg. \emph{k-GAN rel. bg}) it uses the same DCGAN architecture as the object generators, yet maintains its own set of weights. 
It receives as input its own latent sample $\bm{z}_b \sim P(Z_b)$, again using a \emph{UNIFORM(-1, 1)} prior, although one may in theory choose a different distribution.
We explore both variations in which $\bm{z}_b$ participates in the relational stage, and in which it does not. 

\paragraph{Composing} On \emph{Independent MM} and \emph{Triplet MM} we sum the outputs of the object generators as in~\eqref{eq:generator}, followed by clipping to $(0, 1)$.
On all other datasets we use alpha compositing with a fixed order, i.e. using \eqref{eq:alpha_composit}.
In this case the object generators output an additional alpha channel, except for \emph{RGB Occluded MM} in which we obtain alpha values by thresholding the output of each object generator at $0.1$ for simplicity.

\subsection{Hyperparameter Configurations}
Each model is optimized with ADAM~\citep{kingma2015adam} using a learning rate of $0.0001$, and batch size $64$ for $1\,000\,000$ steps.
Each generator step is followed by $5$ discriminator steps, as is considered best practice in training GANs.
Checkpoints are saved at every $20\,000^{th}$ step and we consider only the checkpoint with the lowest FID for each hyper-parameter configuration.
FID is computed using $10\,000$ samples from a hold-out set.

\paragraph{Baseline}
We conduct an extensive grid search over 48 different GAN configurations to obtain a strong GAN baseline on each dataset.
It is made up of hyper-parameter ranges that were found to be successful in training GANs on standard datasets~\citep{kurach2019large}. 

We consider [SN-GAN / WGAN], using [NO / WGAN] gradient penalty with $\lambda$ [1 / 10].
In addition, we consider these configurations [WITH / WITHOUT] spectral normalization.
We consider [(0.5, 0.9) / (0.5, 0.999) / (0.9, 0.999)] as ($\beta_1$, $\beta_2$) in ADAM.
We explore 5 different seeds for each configuration.

\paragraph{k-GAN}
We conduct a similar grid search for the GANs that incorporate our proposed structure. 
However, in order to maintain a similar computational budget compared to our baseline, we consider a \emph{subset} of the previous ranges to compensate for the additional hyper-parameters of the different structured components that we would like to search over.

In particular, we consider SN-GAN with WGAN gradient penalty, with a default $\lambda$ of 1, [WITH / WITHOUT] spectral normalization.
We use (0.9, 0.999) as fixed values for ($\beta_1$, $\beta_2$) in ADAM.
Additionally, we consider K = [3 / 4 / 5] copies of the generator, and the following configurations for the relational structure:

\begin{itemize}
    \item Independent
    \item Relational (1 block, no weight-sharing, 1 head)
    \item Relational (1 block, no weight-sharing, 2 heads)
    \item Relational (2 blocks, no weight-sharing, 1 head)
    \item Relational (2 blocks, weight-sharing, 1 head)
    \item Relational (2 blocks, no weight-sharing, 2 heads)
    \item Relational (2 blocks, weight-sharing, 2 heads)
\end{itemize}

This results in 42 hyper-parameter configurations, for which we each consider 5 seeds.
We do not explore the use of a background generator on the non-background datasets.
On the background datasets, we explore variations with and without the background generator. 
In the former case, we search over an additional hyper-parameter that determines whether the latent representation of the background generator should participate in the relational stage or not, while for the latter we increment all values of K by 1.

\paragraph{IODINE}

We made use of the official trained model for CLEVR released by the authors~\citep{greff2019multi}.
We used $K=7$ to compute the FID, which was also used for training.
Lower values of $K$ were considered but were not found to yield any improvements.

\subsection{Instance Segmentation}

We select the 5 best \emph{3-GAN} models (according to FID) on \emph{Independent MM}, \emph{Triplet MM}, \emph{RGB Occluded MM}, \emph{CIFAR10 + MM}.
These include relational, and purely independent models, although on \emph{CIFAR10 + MM} we ensure that we select only models that also incorporate the background extension. 
Segmentation data is obtained by either thresholding the output of the object generators at 0.1 (on \emph{Independent MM}, \emph{Triplet MM}) or using the (implicit) alpha masks (\emph{RGB Occluded MM}, \emph{CIFAR10 + MM}).
Ground-truth data is obtained in the same fashion, but using the real images of digits and background before combining them.

Our segmentation architecture is similar to the one considered in \cite{chen2019unsupervised} but trained in a supervised fashion.
We use a straight-through estimator that first considers all viable permutations of assigning segmentation outputs to labels, and then back-propagates only the gradients of the permutation for which we measure the lowest cross-entropy loss. 

We test the trained segmenter on ground-truth data and report the Adjusted Rand Index (ARI)~\citep{hubert1985comparing} score on all pixels that correspond to digits in line with prior work~\citep{greff2017neural}.
On \emph{Independent MM}, and \emph{Triplet MM} we additionally ignore overlapping pixels due to ambiguities.

\subsection{Human Study}

We asked human raters to compare the images generated by \emph{k-GAN} $(k=3,4,5)$ to our \emph{GAN} baseline on \emph{RGB Occluded MM}, \emph{CIFAR10 + MM} and \emph{CLEVR}, using the configuration with a background generator for the last two datasets.
For each model, we select the 10 best hyper-parameter configurations, from which we each generate 100 images.
We conduct two different studies 1) in which we compare images from \emph{k-GAN} against \emph{GAN} and 2) in which we asked raters to answer questions about the content (properties) of the images.

\paragraph{Comparison}
We asked reviewers to compare the quality of the generated images.
We asked up to three raters for each image and report the majority vote or “none” if no decision can be reached.
Note that on $CLEVR$ we instructed the raters to ignore visual implausibilities due to floating objects (for both $k-GAN$ and $GAN$) that may arise due to the fixed order in \eqref{eq:alpha_composit}, and measured this effect separately in Figure~\ref{fig:human_eval_properties_clevr}.

\paragraph{Properties}
For each dataset, we asked (up to three raters for each image) the following questions.

On \emph{RGB Occluded MM} we asked:
\begin{enumerate}
\item How many [red, blue, green] shapes are in the image? Answers: [0, 1, 2, 3, 4, 5]
\item How many are recognizable as digits? Answers: [0, 1, 2, 3, 4, 5]
\item Are there exactly 3 digits in the picture, one of them green, one blue and one red? Answers: Yes / No
\end{enumerate}

\noindent On \emph{CIFAR10 + MM} we asked these same questions, and additionally asked:
\begin{enumerate}
\setcounter{enumi}{3}
\item Does the background constitute a realistic scene? Answers: Yes / No
\end{enumerate}

\noindent On \emph{CLEVR} we asked the following set of questions:
\begin{enumerate}
\item How many shapes are in the image? Answers: [0, 1, 2, 3, 4, 5, 6, 7, 8, 9, 10]
\item How many are recognizable as geometric objects? Answers: [0, 1, 2, 3, 4, 5, 6, 7, 8, 9, 10]
\item Are there any objects with mixed colors (eg. part green part red)? Answers: Yes / No
\item Are there any objects with distorted geometric shapes?: Answers: Yes / No 
\item Are there any objects that appear to be floating? Answers: Yes / No
\item Does the scene appear to be crowded? Answers: Yes / No
\end{enumerate}

\clearpage
\section{Overview of Real and Generated Samples}
\label{app:samples}
Generated samples ($8 \times 8$ grid) are shown for the best (lowest FID) structured GAN following our grid search, as well as the best baseline GAN for each dataset.
Real samples from the dataset can also be seen.

\subsection{Independent Multi MNIST}

\begin{figure}[h]
    \centering
    \includegraphics[width=0.9\linewidth]{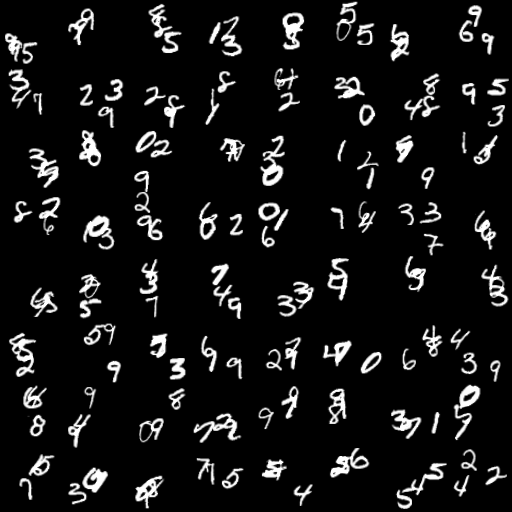}
    \caption{Real}
    \label{fig:real_base_grid}
\end{figure}

\begin{figure}[h]
    \centering
    \includegraphics[width=0.9\linewidth]{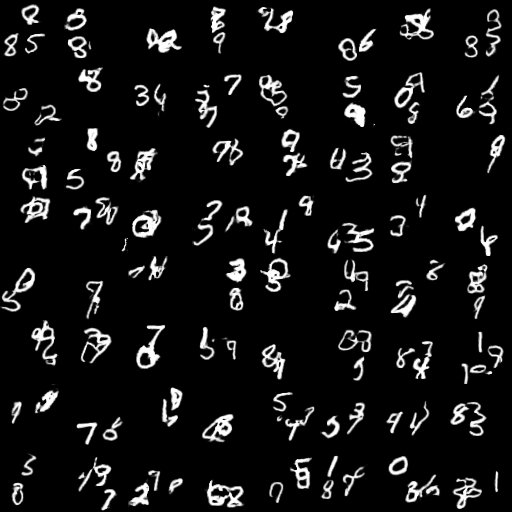}
    \caption{NS-GAN with spectral norm}
    \label{fig:generated_gan_base_grid}
\end{figure}

\begin{figure}[h]
    \centering
    \includegraphics[width=0.9\linewidth]{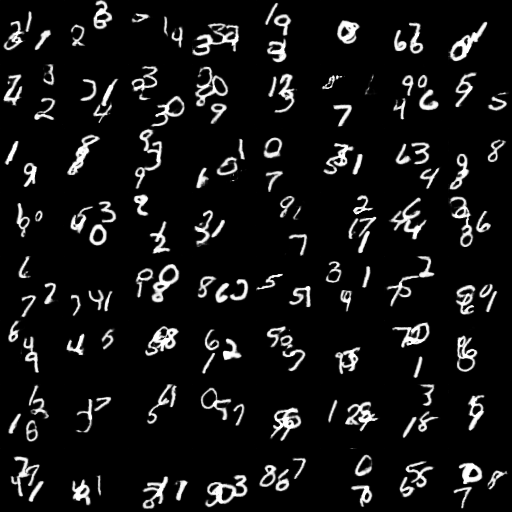}
    \caption{4-GAN ind. with spectral norm and WGAN penalty}
    \label{fig:generated_multi_gan_base_grid}
\end{figure}

Figures~\ref{fig:real_base_grid}--\ref{fig:generated_multi_gan_base_grid}.

\clearpage
\subsection{Triplet Multi MNIST}

\begin{figure}[h]
    \centering
    \includegraphics[width=0.9\linewidth]{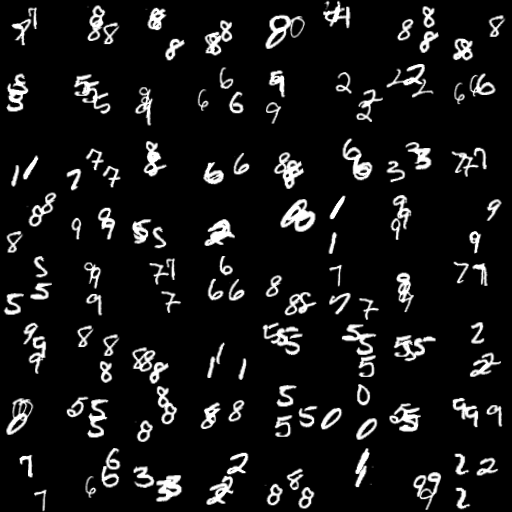}
    \caption{Real}
    \label{fig:real_relational_triplet_grid}
\end{figure}

\begin{figure}[h]
    \centering
    \includegraphics[width=0.9\linewidth]{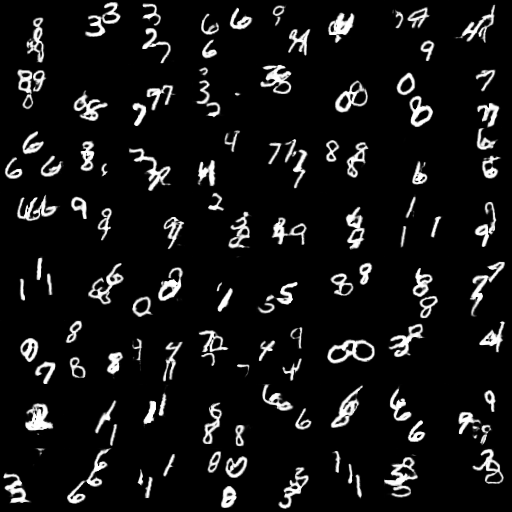}
    \caption{NS-GAN with spectral norm}
    \label{fig:generated_gan_relational_triplet_grid}
\end{figure}

\begin{figure}[h]
    \centering
    \includegraphics[width=0.9\linewidth]{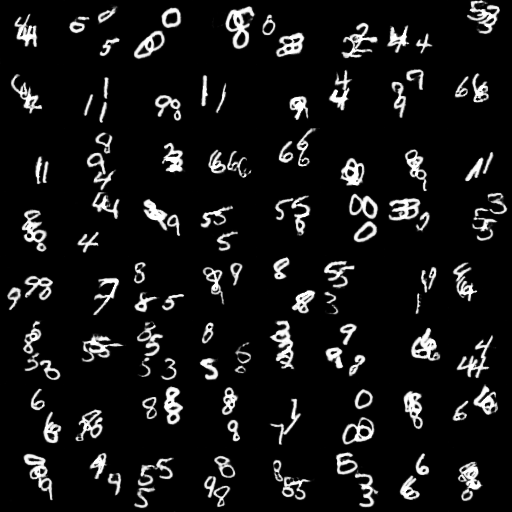}
    \caption{4-GAN rel. (1 block, 2 heads, no weight sharing) with spectral norm and WGAN penalty}
    \label{fig:generated_multi_gan_relational_triplet_grid}
\end{figure}

Figures~\ref{fig:real_relational_triplet_grid}--\ref{fig:generated_multi_gan_relational_triplet_grid}.

\clearpage
\subsection{RGb-Occluded Multi MNIST}

\begin{figure}[h]
    \centering
    \includegraphics[width=0.9\linewidth]{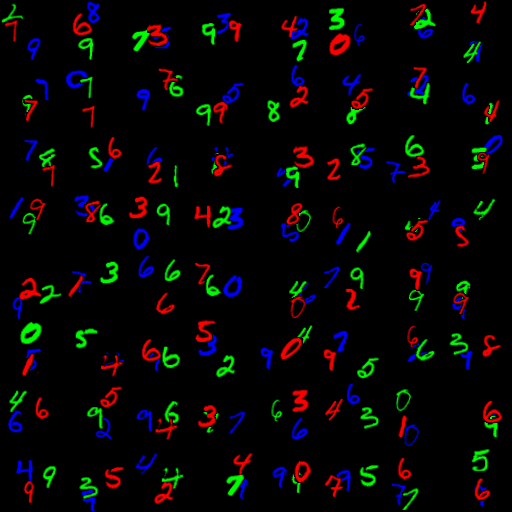}
    \caption{Real}
    \label{fig:real_relational_rgb_occluded_grid}
\end{figure}

\begin{figure}[h]
    \centering
    \includegraphics[width=0.9\linewidth]{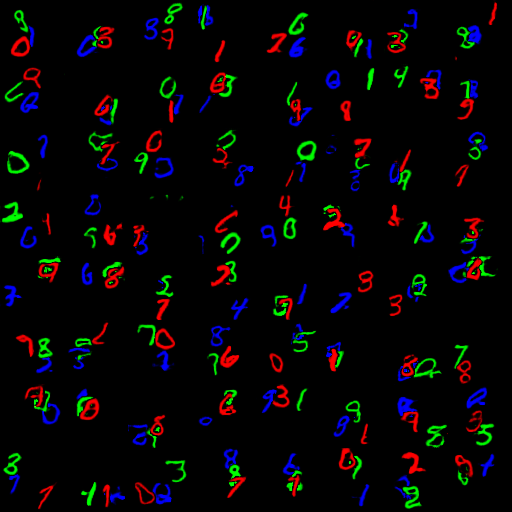}
    \caption{NS-GAN with spectral norm}
    \label{fig:generated_gan_relational_rgb_occluded_grid}
\end{figure}

\begin{figure}[h]
    \centering
    \includegraphics[width=0.9\linewidth]{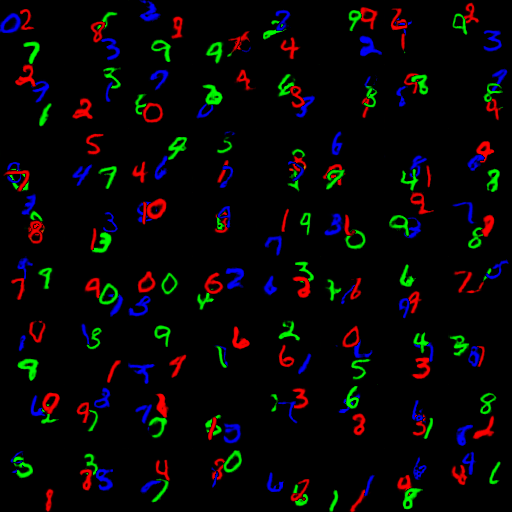}
    \caption{3-GAN rel. (2 blocks, 2 heads, no weight sharing) with spectral norm and WGAN penalty}
    \label{fig:generated_multi_gan_relational_rgb_occluded_grid}
\end{figure}

Figures~\ref{fig:real_relational_rgb_occluded_grid}--\ref{fig:generated_multi_gan_relational_rgb_occluded_grid}.

\clearpage
\subsection{CIFAR10 + MM}

\begin{figure}[h]
    \centering
    \includegraphics[width=0.9\linewidth]{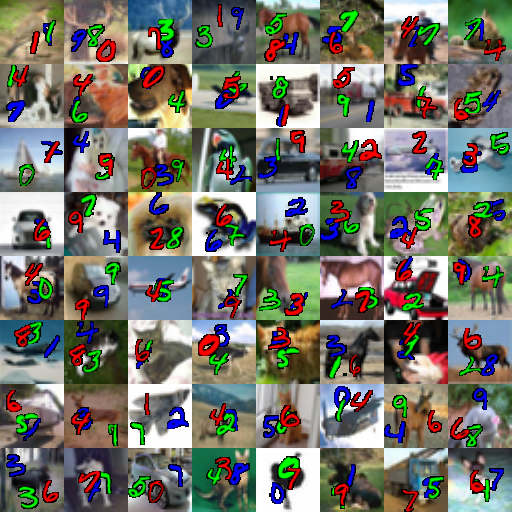}
    \caption{Real}
    \label{fig:real_background_grid}
\end{figure}

\begin{figure}[h]
    \centering
    \includegraphics[width=0.9\linewidth]{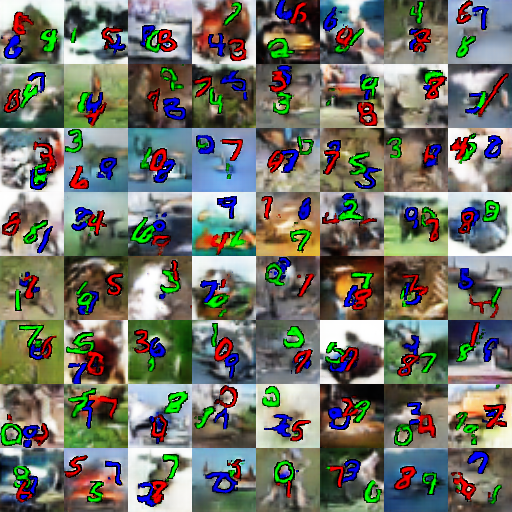}
    \caption{WGAN with WGAN penalty}
    \label{fig:generated_gan_background_grid}
\end{figure}

\begin{figure}[h]
    \centering
    \includegraphics[width=0.9\linewidth]{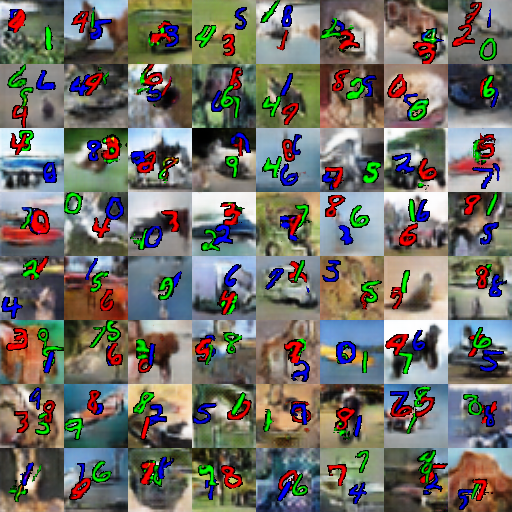}
    \caption{5-GAN rel. (1 block, 2 heads, no weight sharing) bg. (no interaction) with WGAN penalty}
    \label{fig:generated_multi_gan_bg_background_grid}
\end{figure}

Figures~\ref{fig:real_background_grid}--\ref{fig:generated_multi_gan_bg_background_grid}.

\clearpage
\subsection{CLEVR}

\begin{figure}[h]
    \centering
    \includegraphics[width=0.9\linewidth]{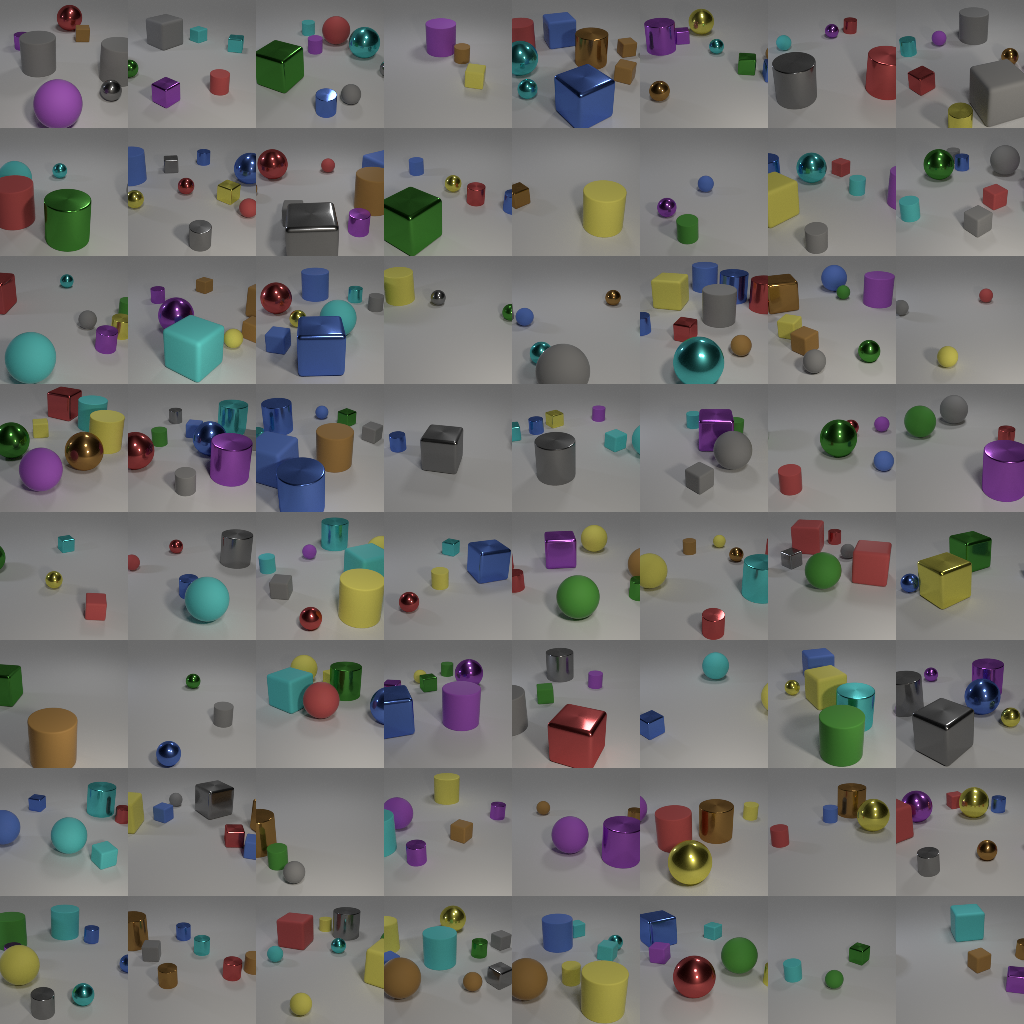}
    \caption{Real}
    \label{fig:real_clevr_grid}
\end{figure}

\begin{figure}[h]
    \centering
    \includegraphics[width=0.9\linewidth]{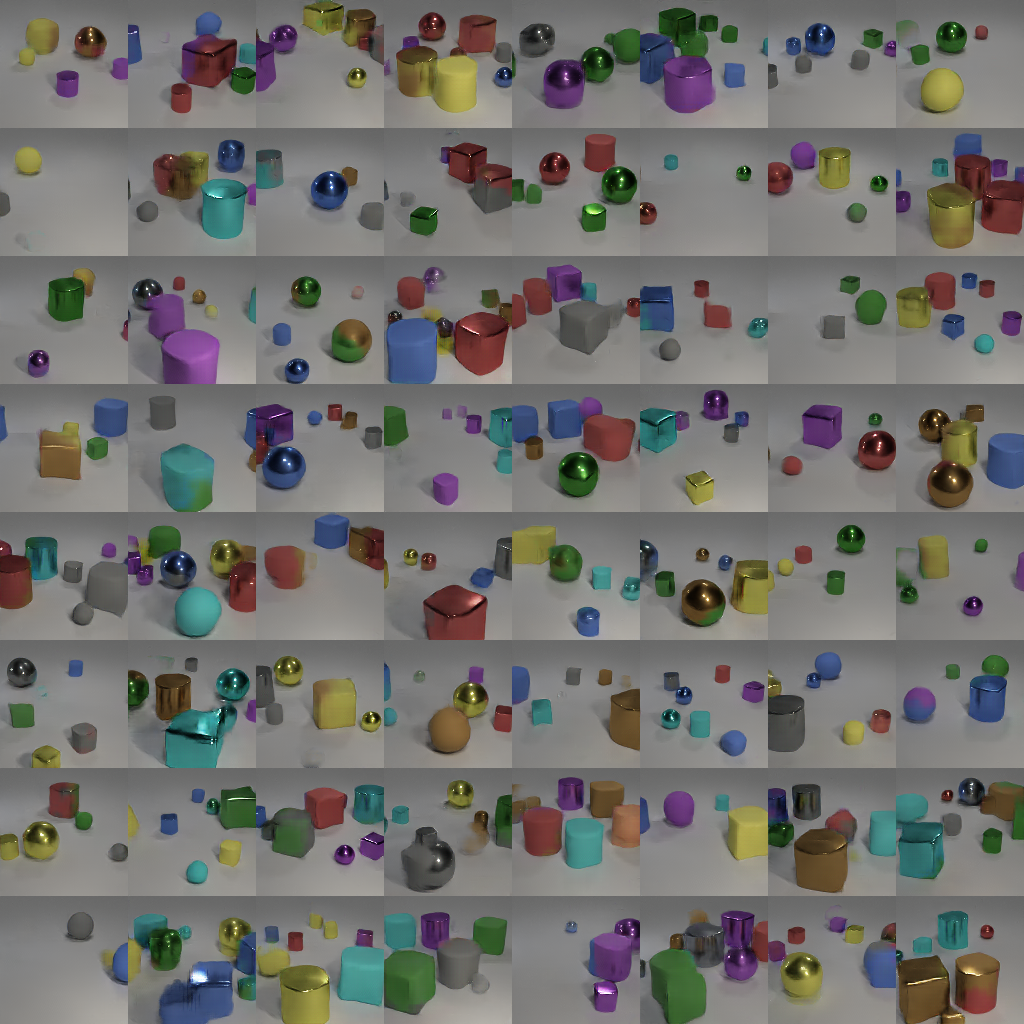}
    \caption{WGAN with WGAN penalty}
    \label{fig:generated_gan_clevr_grid}
\end{figure}

\begin{figure}[h]
    \centering
    \includegraphics[width=0.9\linewidth]{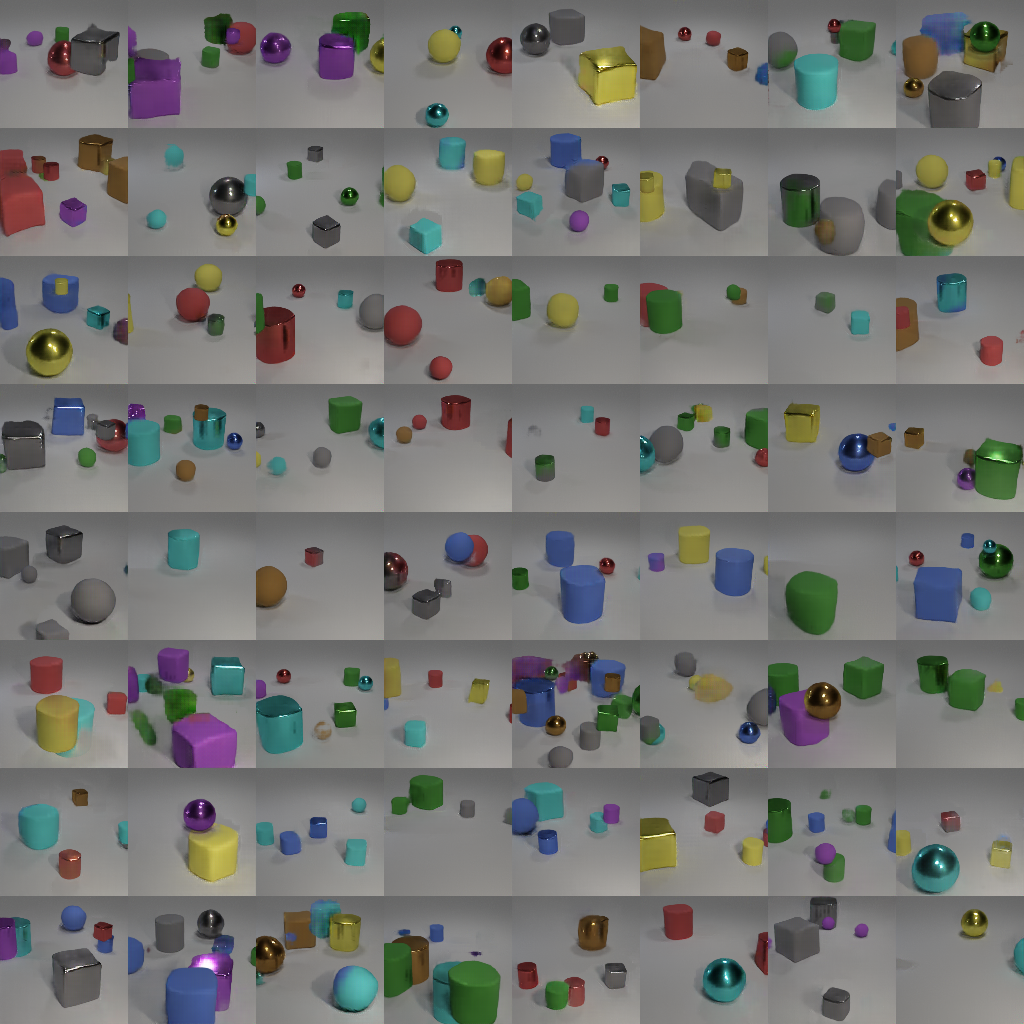}
    \caption{3-GAN rel. (2 heads, 2 blocks, no weight sharing) bg. (with interaction) with WGAN penalty.}
    \label{fig:generated_multi_gan_bg_clevr_grid}
\end{figure}

Figures~\ref{fig:real_clevr_grid}--\ref{fig:generated_multi_gan_bg_clevr_grid}.

\end{document}